\documentclass[times,twocolumn,final]{elsarticle}
\usepackage{medima}
\usepackage{framed,multirow}

\usepackage{amssymb}

\usepackage{url}
\usepackage{xcolor}
\usepackage{hyperref}

\usepackage{latexsym}
\usepackage{graphicx}
\usepackage{kotex}
\usepackage{booktabs}
\usepackage{array}
\usepackage{makecell}
\usepackage{nicefrac}  
\usepackage{microtype} 
\usepackage{mathtools}
\usepackage{amsfonts}
\usepackage{dsfont}
\usepackage{cuted}
\usepackage{graphicx}
\usepackage{kotex}
\usepackage{multirow}
\usepackage{booktabs}
\usepackage{array}
\usepackage{makecell}
\usepackage{amsmath}
\usepackage{xspace}
\usepackage{cases}
\usepackage{pifont}
\usepackage{breqn}
\usepackage{caption} 
\usepackage{enumitem}
\usepackage{algpseudocode}

\usepackage[linesnumbered,ruled]{algorithm2e}

\newcommand{\cmark}{\ding{51}}%
\newcommand{\xmark}{\ding{55}}%
\DeclareMathOperator{\E}{\mathbb{E}}
\newcommand{\cvm}{{CVM}\xspace}
\newcommand{\ourdata}{{Lateral cephalometric radiograph}\xspace}
\newcommand{\spinewebdata}{AASCE\xspace}
\newcommand{\buudataAP}{BUU-AP\xspace}
\newcommand{\buudataLA}{BUU-LA\xspace}

\newcommand{\ourmethod}{{Attend-and-Refine Network}\xspace}
\newcommand{\ourshort}{{ARNet}\xspace}
\newcommand{\attendOurs}{{Interaction-guided recalibration network}\xspace}
\newcommand{\blue}[1]{\textcolor{black}{#1}}
\newcommand{\rbt}[1]{\textcolor{black}{#1}}
\newcommand{\rbttwo}[1]{\textcolor{black}{#1}}
\newcommand{\rbtthree}[1]{\textcolor{black}{#1}}

\definecolor{newcolor}{rgb}{.8,.349,.1}

\journal{Medical Image Analysis}

\begin{document}

\verso{Jinhee Kim \textit{et~al.}}

\begin{frontmatter}

\title{Attend-and-Refine: Interactive keypoint estimation and quantitative cervical vertebrae analysis for bone age assessment} 
% \tnoteref{tnote1}}%
% \tnotetext[tnote1]{This is an example for title footnote coding.}

\author[1]{Jinhee \snm{Kim}}
\author[1]{Taesung \snm{Kim}} %\fnref{fn1}}
% \fntext[fn1]{This is author footnote for second author.}
\author[1]{Taewoo  \snm{Kim}}
\author[2]{Dong-Wook \snm{Kim}}
\author[3]{Byungduk \snm{Ahn}}
\author[4]{Yoon-Ji \snm{Kim}\corref{cor1}}
\ead{yn0331@ulsan.ac.kr}
\author[2]{In-Seok \snm{Song}\corref{cor1}}
\ead{densis@korea.ac.kr}
%% Third author's email
\author[1]{Jaegul \snm{Choo}\corref{cor1}}
\ead{jchoo@kaist.ac.kr}
\cortext[cor1]{Corresponding authors: Tel.: +82-42-350-1813 (Jaegul Choo);}
\address[1]{Kim Jaechul Graduate School of Artificial Intelligence, KAIST, Daejeon 34141, Republic of Korea}
\address[2]{Korea University Anam Hospital, Seoul 02708, Republic of Korea}
\address[3]{Papa’s Dental Clinic, Seoul 06593, Republic of Korea}
\address[4]{Asan Medical Center, University of Ulsan College of Medicine, Seoul 05505, Republic of Korea}

\received{}
\finalform{}
\accepted{}
\availableonline{}
\communicated{}

\begin{abstract}
In pediatric orthodontics, accurate estimation of growth potential is essential for developing effective treatment strategies. Our research aims to predict this potential  by identifying the growth peak and analyzing cervical vertebra morphology solely through lateral cephalometric radiographs. We accomplish this by comprehensively analyzing cervical vertebral maturation (CVM) features from these radiographs. This methodology provides clinicians with a reliable and efficient tool to determine the optimal timings for orthodontic interventions, ultimately enhancing patient outcomes. A crucial aspect of this approach is the meticulous annotation of keypoints on the cervical vertebrae, a task often challenged by its labor-intensive nature. To mitigate this, we introduce Attend-and-Refine Network (ARNet), a user-interactive, deep learning-based model designed to streamline the annotation process. ARNet features Interaction-guided recalibration network, which adaptively recalibrates image features in response to user feedback, coupled with a morphology-aware loss function that preserves the structural consistency of keypoints. This novel approach substantially reduces manual effort in keypoint identification, thereby enhancing the efficiency and accuracy of the process. Extensively validated across various datasets, ARNet demonstrates remarkable performance and exhibits wide-ranging applicability in medical imaging. In conclusion, our research offers an effective AI-assisted diagnostic tool for assessing growth potential in pediatric orthodontics, marking a significant advancement in the field.
\end{abstract}


\end{frontmatter}

\begin{figure*}[t!]
\begin{center}
  \includegraphics[width=0.9\linewidth]{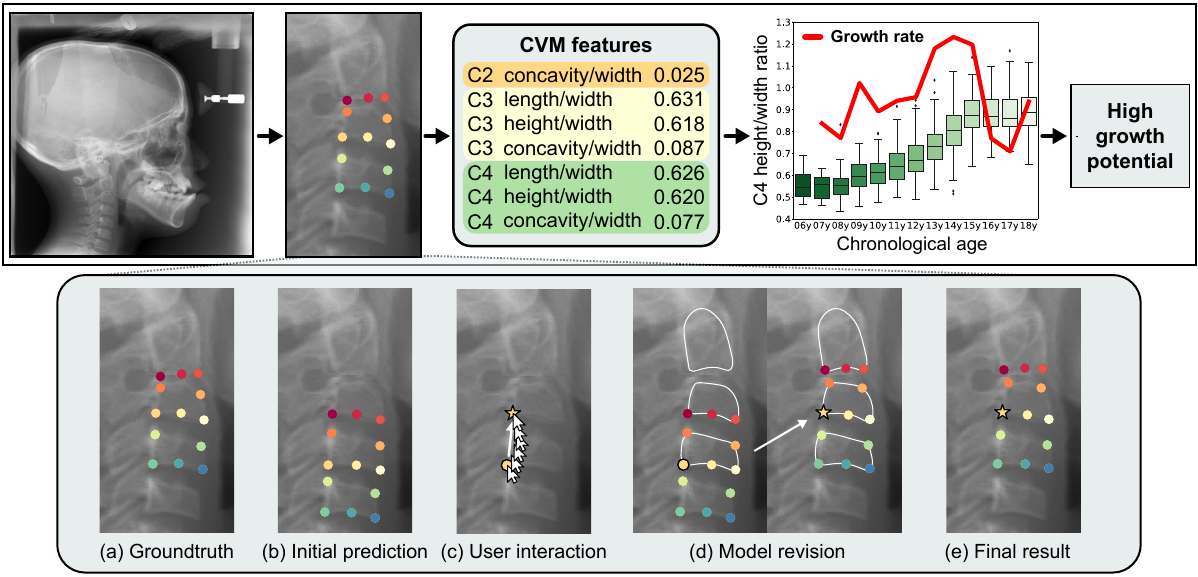}
\end{center}
\vspace{-10pt}
  \caption{\textbf{Growth potential assessment via interactive keypoint estimation.} Our method significantly enhances accuracy and efficiency of growth potential assessment in pediatric orthodontics. The process begins with the initial keypoint estimation phase, where the objective is to identify (a) the keypoints on the second, third, and fourth cervical vertebrae. However, the model occasionally fails to detect these keypoints accurately, (b) missing the second vertebrae at the top in this example. Conventionally, rectifying such an error would necessitate manually revising all 13 keypoints, akin to starting the annotation from scratch. However, our method introduces a substantial improvement: a user needs to correct only (c) one keypoint, which automatically prompts (d) the adjustment of the remaining keypoints. Consequently, this leads to (e) a final output that accurately reflects the user's interaction as well as correctly detects the target keypoints, thereby demonstrating the accuracy and the efficiency of our approach. 
The keypoints identified through this process are then utilized to analyze the cervical vertebrae morphology, quantified as cervical vertebral maturation (\cvm) features. These features enable us to assess the patient's growth potential using standard growth curves. } 
\label{fig:pipeline}
\end{figure*}

\section{Introduction}
\label{sec1} 

Accurate prediction of growth potential during childhood and adolescence holds significant value in orthodontic diagnosis and treatment planning~\citep{med_4,cvm2}. For instance, treating conditions such as prognathism before the growth peak may lead to relapse, or addressing underdeveloped jaws after the growth peak might prove ineffective. Thus, accurately identifying the growth peak is essential for determining the most effective treatment period for orthodontic conditions in children.
Numerous studies have demonstrated a reliable association between skeletal maturity and the developmental potential of various body parts, such as the jaw and statural height~\citep{g3,g2,g1,g5,g7,g4,g6}. 
Traditionally, bone age assessment using hand-wrist radiographs has been the standard for evaluating skeletal maturity~\citep{med_5,med_2,cvm3,med_1,med_3}.
The cervical vertebral maturation (\cvm) method has gained attention in clinical orthodontic practice~\citep{mito2002cervical,med_8,cvm3,cvm2}.
Consequently, using CVM for bone age assessment prevents additional radiographic exposure to children and adolescents who comprise the majority of orthodontic patients. 
Studies have demonstrated a correlation between \cvm stages and significant developmental phases, such as mandible growth spurts or changes in statural height~\citep{g9,baccetti2002improved,hosni2018comparison}, and their relationship with chronological age~\citep{mito2002cervical,med_8,med_1,med_4}.
 These studies underscore CVM as a reliable indicator of jaw growth and development.
 
Despite these insights, previous research has often relied on qualitative assessments of cervical vertebrae morphology with limited data. 
Therefore, our study aims to undertake a comprehensive quantitative analysis of cervical vertebrae morphology to estimate the pubertal growth peak more accurately during childhood and adolescence. 
By analyzing extensive datasets, our objective is to assess the annual growth rate of the cervical vertebrae to pinpoint the period of growth peak and the corresponding cervical vertebrae morphology.
Our approach, leveraging morphological characteristics at growth peak as a predictor of a patient's growth potential, offers detailed diagnostic guidance and aids clinicians in determining the optimal timing for orthodontic treatments.

A vital component of this quantitative analysis is the precise annotation of keypoints on the cervical vertebrae, which can be a labor-intensive task when done manually.
Automated methods, though intended to reduce manual effort, often need more accuracy and still require clinician validation, thus becoming as burdensome as manual annotation~\citep{kim2022morphology}.
To address these challenges, we introduce \textbf{\ourmethod} (\textbf{\ourshort}), a novel {interactive keypoint estimation framework} designed to accurately estimate keypoints through effective user interaction. By integrating automated predictions with user expertise, this methodology significantly enhances both accuracy and efficiency in the annotation process.

\rbt{\rbtthree{Up to our knowledge, there has been limited research on interactive keypoint estimation frameworks. Interactive image segmentation, a widely studied field, shares conceptual similarities with interactive keypoint estimation by integrating user feedback to enhance accuracy and can serve as a valuable foundation for developing interactive keypoint estimation models. Interactive segmentation models, such as BRS~\citep{brs}, f-BRS~\citep{fbrs}, RTIM~\citep{ritm}, have significantly advanced precision and efficiency. }}

\rbt{However, these models exhibit notable limitations when adapted to our proposed interactive keypoint estimation framework. User interaction information often impacts only the local area around the user correction, failing to propagate corrections to distant keypoints. In some cases, model revision even degrades initial accuracy (Refer to Fig.~\ref{suppe_fig:qual} of Appendix~\ref{appen_sec:qual}).}

\rbt{These limitations arise from the distinct challenges of interactive segmentation and keypoint estimation. For example, interactive segmentation focuses on segmenting adjacent pixels near the user input, while interactive keypoint estimation requires propagating corrections to non-connected, distant keypoints. This underscores the need for a specialized approach to interactive keypoint estimation.}

\ourshort addresses these challenges via \textbf{\attendOurs}, a mechanism designed to recalibrate image features using user modification information, ensuring alignment with user feedback, and \textbf{morphology-aware loss function}, which maintains consistent inter-keypoint relationships and ensures that related keypoints are updated correspondingly when user corrects a keypoint.
\textbf{\ourshort-v1}, proposed in our previous work~\citep{kim2022morphology}, utilizes global pooling and squeeze-and-excitation layers~\citep{squeezeAndExcitation} to recalibrate backbone image features channel-wise based on user interaction signals, enabling the propagation of user feedback to distant keypoints. 

\begin{figure}[!t]
\begin{center}
\includegraphics[width=1.0\linewidth]{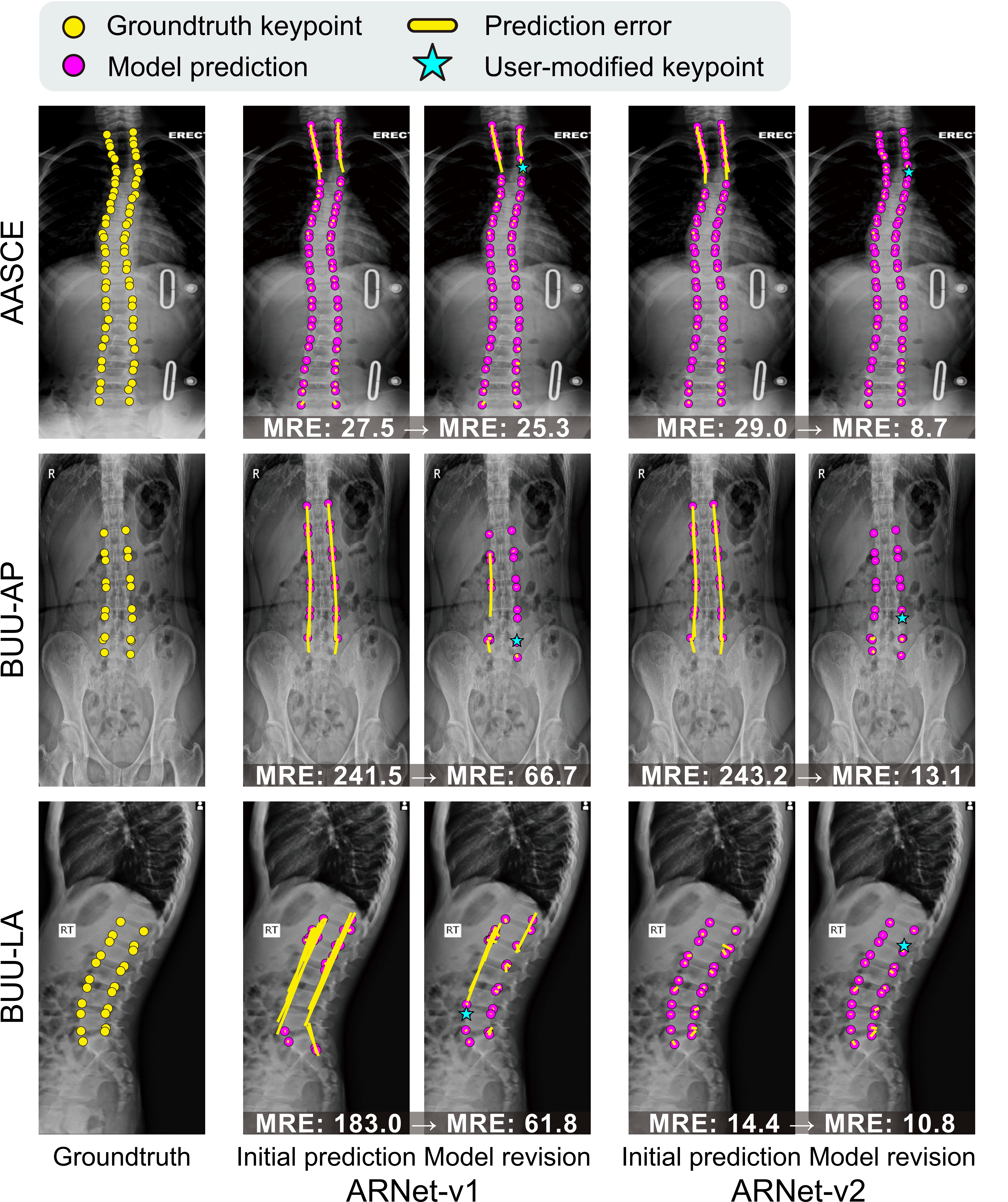}
\end{center}
\vspace{-5pt}
  \caption{\rbt{\textbf{Prediction results of ARNet-v1 and ARNet-v2 on the public datasets.} \textit{Model revision} represents the updated outputs from each model after incorporating a single user correction into the initial predictions. The mean radial error (MRE) for each prediction is denoted.}
}
\label{fig:Arnet2_intro}
\end{figure}

\rbt{\rbtthree{However, ARNet-v1 applies interaction signals uniformly across spatial locations, lacking pixel-level specificity. This often results in significant residual errors, particularly in challenging cases, as shown in Fig.~\ref{fig:Arnet2_intro}. 
To overcome these limitations, \textbf{ARNet-v2} introduces a cross-attention-based mechanism~\citep{chefer2023attend} that selectively integrates user interaction signals at the pixel level. Image features serve as queries, while user interaction signals act as keys and values, enabling globally aware, pixel-specific retrieval of relevant information. This design ensures the adaptive propagation of user interaction signals to all keypoints, leading to precise corrections. 
As shown in Fig.~\ref{fig:Arnet2_intro}, ARNet-v2 significantly reduces errors across all incorrect keypoints with a single user correction and achieves substantially lower initial prediction errors, outperforming ARNet-v1.
}}

\rbt{\rbtthree{In summary, ARNet-v2 leverages a cross-attention-based mechanism to enable globally aware and selective retrieval of user interaction signals, effectively propagating corrections to all relevant keypoints. This approach leads to superior performance, significantly enhancing both the accuracy and efficiency of keypoint estimation.
Extensive experiments demonstrate that ARNet-v2 consistently achieves state-of-the-art results across four datasets, outperforming both ARNet-v1 and baseline models. For example, on the AASCE dataset, ARNet-v2 reduces the failure rate by 37\% compared to ARNet-v1 and by 67\% compared to the closest baseline. Additionally, ARNet-v2 also decreases the number of clicks by 25\% compared to ARNet-v1 and by 43\% compared to the closest baseline. These findings highlight its effectiveness in delivering accurate keypoint estimation with minimal user input.
}}

Additionally, our research presents a web-based AI-assisted growth potential estimation tool, as illustrated in Fig.~\ref{fig:pipeline}.
 A supplementary video further demonstrates the practical application of our method, highlighting its effectiveness. This study provides clinicians with a practical and easy-to-use diagnostic guideline for determining optimal timing for orthodontic treatments, presenting a significant advancement in the field.

\section{Related work}
\subsection{Skeletal maturity assessment}\label{sec:relatedwork_skeletal_mautrity}            
The assessment of skeletal maturity using hand-wrist bone radiographs is a pivotal area in pediatric radiology and orthopedics. This field has been significantly advanced by the methodologies developed by Greulich and Pyle~\citep{greulich1959radiographic}, Tanner and Whitehouse~\citep{tanner1976clinical}, and Fishman~\citep{fishman1982radiographic}, contributed unique approaches to hand-wrist bone age assessment. Concurrently, the cervical vertebral maturation method, initially introduced by Lamparski~\citep{Lamparski1975}, and later advanced by Hassel and Farman~\citep{hassel1995skeletal}, offers an alternative approach. This method focuses on cervical vertebrae morphological changes as indicators of skeletal growth stages~\citep{szemraj2018cervical}. 

The correlation between CVM stages and growth spurts has been well-documented in various studies~\citep{g9,mito2002cervical}.
Notably, Tancan et al.~\citep{med_8} highlights the clinical significance of \cvm stages as markers of the pubertal growth period, linked closely with chronological age. 
Baccetti et al.~\citep{baccetti2002improved} refined the CVM  method by analyzing the morphology of the second, third, and fourth cervical vertebrae in the lateral cephalograms and  relating to the growth stages of the mandible. 
The effectiveness of the approach is supported by numerous studies demonstrating a significant correlation between skeletal maturation observed in hand-wrist bones and cervical vertebrae~\citep{mito2002cervical,alkhal2008correlation,szemraj2018cervical}. 
Kim et al.~\citep{kim2021prediction} have developed a machine learning-based approach for assessing bone age based on CVM features, marking a significant contribution in increasing accuracy and efficiency of CVM assessments. 

Building upon this, our work aims to predict the timing of pubertal growth peak by utilizing CVM features derived from lateral cephalometric radiographs, employing a deep learning-based approach.
This approach sets itself apart from previous work by focusing on a quantitative analysis of cervical vertebrae morphology to identify the pubertal growth peak across a substantial dataset. Our method offers a higher level of precision compared to earlier methods that predominantly relied on graphical representation.  
By accurately pinpointing the pubertal growth peak and the corresponding cervical vertebrae morphology, our study enhances the accuracy of growth potential assessments and enriches the understanding of cervical vertebral development during critical stages of childhood and adolescence. Our work, therefore, paves the way for a  precise and cost-effective approach in dentofacial orthopedic treatment planning, offering deeper insights and more reliable tools for clinicians.

\begin{figure*}[t!]
\begin{center}
  \includegraphics[width=1.0\linewidth]{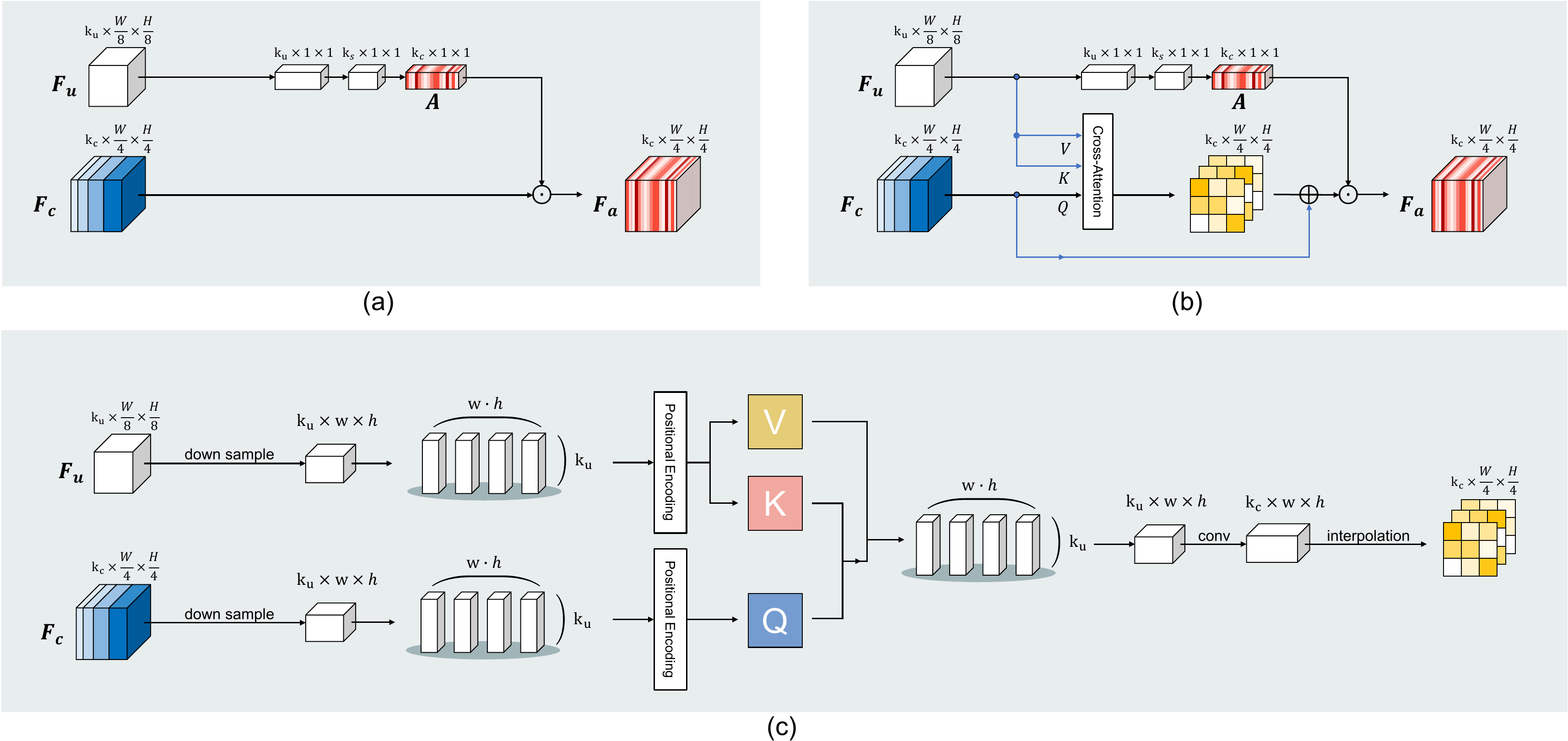}
\end{center}
\vspace{-10pt}
  \caption{{\textbf{\attendOurs}. (a) \ourshort-v1. (b) \ourshort-v2. (c) Details of the cross-attention layer in \ourshort-v2. \rbt{User interaction information is used as value (\textit{V}) and key (\textit{K}), and image features are used as query (\textit{Q}). Notations are summarized in Table~\ref{table:notation} in Appendix~\ref{appen_sec:notation}.}}  
}
\label{fig:arnet}
\end{figure*}

\subsection{Keypoint detection in medical domain}\label{sec:relatedwork_keypoint_detection}
Keypoint detection is a critical task in medical image analysis and has traditionally been dependent on manual annotation by clinicians~\citep{manual2,manual1}. While effective, this manual approach is time-consuming and labor-intensive, which limits its practicality for large-scale analysis.
Recently, there has been extensive research on deep learning-based methods for automating keypoint detection in medical images, achieving notable results~\citep{cephann,recent0,recent1,kim2022morphology,yuan2021hrformer,kim2024bones}. 
However, a significant challenge that these developments face is the considerable human effort required to verify and review model predictions, which is crucial in medical fields where accuracy is of utmost importance.
To address this issue, our study proposes an interactive estimation method. This innovative approach is designed to reduce the extensive effort involved in reviewing model predictions, aiming to substantially enhance both the accuracy and efficiency of keypoint detection in medical imaging. Such an advancement is especially valuable in orthodontics and related medical areas, where precise keypoint-based analysis is vital for effective patient care.

\subsection{Interactive image segmentation}\label{sec:relatedwork_interactive_segmentation}
Interactive image segmentation models share a common goal with interactive keypoint estimation, which is to integrate user feedback into model predictions. 
As such, they can provide a strong foundation for the development of an interactive keypoint estimation framework.
Existing studies have significantly improved precision and efficiency in the field of image segmentation. Backpropagation refinement methods, such as BRS~\citep{brs} and f-BRS~\citep{fbrs}, have demonstrated remarkable effectiveness in refining segmentation accuracy. Notable advancements include FCA-Net~\citep{firstclick}, which emphasizes the importance of a user’s initial click and has achieved high accuracy on datasets like MSCOCO. Similarly, RITM~\citep{ritm} employs an iterative training approach, leveraging prediction masks from previous iterations to enhance performance stability.

However, these methods face limitations when applied to keypoint detection. The fundamental difference lies in the nature of input and expected output.
Interactive image segmentation models typically process \textit{interior} or \textit{boundary} points to segment the main body of a target object. In contrast, our interactive keypoint estimation model starts with predefined keypoints and receives the groundtruth location for one keypoint, adjusting others that are often in \textit{non-connected}, \textit{non-adjacent} areas. Moreover, unlike interactive image segmentation, where initial user input is necessary, interactive keypoint estimation starts with a model-generated prediction of predefined keypoints without user input, which is then iteratively refined based on user feedback.

To effectively tackle the unique challenges of interactive keypoint estimation, we propose a novel architecture specifically tailored for this task. A critical feature that sets our model apart from interactive image segmentation models is the incorporation of morphology-aware loss function. This function is designed to maintain morphological consistency among keypoints, ensuring updates to related keypoints in a cohesive manner. This feature is vital for interactive keypoint estimation, as it enhances the accuracy and coherence of adjustments based on user interactions.

\subsection{Our work}
Our approach introduces a novel interactive keypoint estimation framework, providing an efficient alternative to traditional manual annotation by automatically refining errors based on user corrections.
This paper builds on our previous work~\citep{kim2022morphology}, which, to our knowledge, proposed the \textit{first} interactive keypoint estimation model, named ARNet-v1, in this study. 
This work introduces significant advancements, including (1)
ARNet-v2, which addresses key limitations of ARNet-v1 with new technical contributions, (2) extensive experiments conducted on two more public datasets, (3) growth peak analysis, adding practical contributions for orthodontic treatment planning.

First, \rbt{\rbtthree{ARNet-v2 represents a substantial advancement over ARNet-v1 by overcoming its limitations related to uniform scaling. By leveraging a cross-attention-based mechanism, ARNet-v2 enables globally aware and selective retrieval of user interaction signals, effectively propagating this information to all relevant keypoints. This approach significantly enhances the accuracy and efficiency of keypoint estimation and achieves superior performance compared to baseline models.}}

Second, the robustness and versatility of ARNet-v2 are comprehensively validated through extensive experiments on additional public datasets, demonstrating its outstanding performance and adaptability to diverse medical imaging scenarios. These evaluations underscore the suitability of the proposed model for real-world clinical applications.

Third, \rbt{\rbtthree{our growth peak analysis using cervical vertebral keypoints adds a valuable practical dimension, providing effective tools for orthodontic treatment planning. We quantitatively analyze cervical vertebrae morphology during growth phases and leverage morphological characteristics at growth peak as a predictor of a patient’s growth potential. This analysis provides valuable clinical insights and demonstrates the practical use of our model in clinical settings. Our approach significantly improves clinical decision-making processes by assisting in timely and appropriate medical interventions.}}

These contributions underscore the substantial improvements over our previous work. By addressing limitations and expanding the practical applications of interactive keypoint estimation, this study advances the field of medical image analysis.

\section{\ourmethod}\label{sec:keypoint}
We introduce \ourshort, a sophisticated method for interactive keypoint estimation in radiographs (Section~\ref{sec:keypointoverview}), integrating \attendOurs (Section~\ref{sec:interaction}) and a morphology-aware loss (Section~\ref{sec:morph}).
\ourshort is designed to significantly improve the efficiency of keypoint annotation by achieving improved accuracy in initial prediction and streamlining the keypoint refinement process. Initially, the model independently identifies keypoints within an image. Users then review and, if necessary, adjust these keypoints to enhance their accuracy.
A critical feature of \ourshort is its capability to automatically adjust associated mispredicted keypoints based on minimal user corrections. This functionality is crucial in reducing manual effort, as it lessens the need for individual corrections of each keypoint.
Such an advancement is especially valuable in clinical environments where precision and efficiency are paramount.

\subsection{Interactive keypoint estimation methodology}\label{sec:keypointoverview}
Consider a radiographic image $\mathcal{I}\in \mathds{R}^{C \times W \times H}$, where $C$, $W$, and $H$ represent the number of channels, width, and height, respectively. 
\ourshort processes this image to generate heatmaps $\hat{\mathcal{H}} \in \mathds{R}^{K \times W \times H}$ for $K$ designated keypoints.
The training process of \ourshort incorporates a dual-phase approach, beginning with the initial prediction of keypoints from the radiograph, followed by a refinement phase that incorporates simulated user inputs. This iterative process enables \ourshort to not only make initial accurate keypoint predictions but also dynamically refine these predictions in response to user feedback. 

\subsubsection{Simulation of user interaction}
 \ourshort simulates user interactions to refine model predictions of keypoints, utilizing an iterative training approach in line with clinical scenarios where users repeatedly revise keypoint predictions. 
 The process begins by determining the number of modifications to apply to a model prediction. This number is selected from a multinomial distribution within the range of $[0, K]$. The distribution is skewed towards fewer modifications, aligning with our goal of enhancing efficiency.
Following this, keypoints for correction are randomly selected using a discrete uniform distribution, $\text{Unif}(1, K)$, ensuring an equal probability of selection for each keypoint.
These selected keypoints are subsequently encoded into heatmaps using groundtruth coordinates, effectively simulating user interactions. 

\subsubsection{User interaction encoding}
User modifications are represented by $\mathcal{U} \in \mathds{R}^{K \times W\times H}$ .
When a user adjusts $k$ keypoints, denoted by $\{c_1, c_2, ... c_k\}$, the corresponding channels $\{\mathcal{U}_{c_1}, \mathcal{U}_{c_2}, ..., \mathcal{U}_{c_k}\}$ are encoded as Gaussian-smoothed heatmaps using the updated user-provided coordinates. Channels corresponding to unmodified keypoints are set to zero. 
The user interaction heatmap for the $n$-th keypoint is defined as: 
\begin{equation}
  \mathcal{U}_n(i,j) = \begin{cases}
     \exp\Big(\frac{(i-x_n)^2+(j-y_n)^2}{-2\sigma_u^2}\Big), & \text{if $n\in \{c_1, c_2, ..., c_k\}$}.\\
    0, & \text{otherwise},
  \end{cases}
\end{equation}
where $\sigma_u^2$ is the pre-determined variance. Here, $p_n=(x_n, y_n)$ refers to the user-modified coordinates of the $n$-th keypoint. In training and testing, this is set as ground-truth keypoint coordinates, while in clinical applications, it represents the actual coordinates as provided by clinicians. 

\subsubsection{Keypoint coordinate encoding}
The model is trained to estimate the heatmaps of keypoint locations using the binary cross-entropy loss between the prediction and the groundtruth heatmaps. 
The groundtruth x-y coordinates of the target keypoints are transformed into Gaussian-smoothed heatmaps $\mathcal{H} \in \mathds{R}^{K \times W \times H}$, following methodologies from previous studies~\citep{K11,hrnetv2}.
For each keypoint $n$ with groundtruth coordinates $p_n=(x_n, y_n)$, the corresponding groundtruth heatmap is generated as:
\begin{equation}
  \mathcal{H}_n(i,j) = \exp\Bigg(\frac{(i-x_n)^2+(j-y_n)^2}{-2\sigma_h^2}\Bigg),
    \label{eq:heatmap}
\end{equation}
where $\sigma_h^2$ denotes the variance used for Gaussian, and $i$ and $j$ represent the pixel coordinates.

\subsubsection{Keypoint coordinate decoding.}
To determine the final keypoint estimates, the predicted heatmaps $\hat{\mathcal{H}} \in \mathds{R}^{K \times W \times H}$ are converted into 2D keypoint coordinates.
This is accomplished using a differentiable local soft-argmax method, which reduces quantization errors inherent in heatmap formats~\citep{subpixel}.
The process begins by identifying the highest probability coordinate for each keypoint, denoted as $\hat{q}_n =\text{argmax } \hat{\mathcal{H}}_n$.
A patch around this coordinate is then extracted, to which the soft-argmax operation is applied to compute an offset. Finally, the accurate keypoint coordinates $\hat{p_n}$ are obtained by adding this offset to the initially identified coordinates $\hat{q}_n$, leading to more precise keypoint localization.

\subsection{\attendOurs}\label{sec:interaction}
\ourshort enhances the keypoint refinement process by leveraging user feedback, which serves as a guide to adjust mispredicted keypoints.
\attendOurs plays a pivotal role in this process by interpreting user interactions to refine keypoint localization.
It works by creating feature maps that encapsulate the specifics of these interactions and transforming the original image features into user-feedback-adjusted keypoint location heatmaps.
This ensures that the feature map aligns precisely with user feedback, leading to the generation of accurate keypoint heatmaps.
The process begins by merging user interaction heatmaps $\mathcal{U}$ with a downsampled feature map $\mathbf{F}_i$ from the main network. This composite data is then encoded into lower resolution features, $\mathbf{F}_u$. \attendOurs extracts critical user interaction information from $\mathbf{F}_u$ to recalibrate the original image feature map $\mathbf{F}_c$.

We introduce two versions of the network: \ourshort-v1 and \ourshort-v2, as depicted in Fig.~\ref{fig:arnet}.
\textbf{\ourshort-v1} (Fig.~\ref{fig:arnet}a) employs a gating mechanism with a squeeze-and-excite layer~\citep{squeezeAndExcitation}, focusing on adaptive recalibration according to channel-wise feature responses.
It employs global max-pooling on $\mathbf{F}_u$ to capture the most salient responses, reducing its spatial resolution to $1\times1$.
These are then processed through fully connected layers and a sigmoid activation function to produce gating weights $A$. Finally, these weights are multiplied channel-wise to the original feature map $\mathbf{F}_c$, enabling channel-wise modulation of the image feature maps in response to user interactions. 

\textbf{\ourshort-v2} (Fig.~\ref{fig:arnet}b-c) addresses the limitations of \ourshort-v1, particularly in scenarios requiring fine adjustments over larger areas.
It incorporates a cross-attention layer known for its effectiveness in user input-driven controllable image generation~\citep{rombach2022high,chefer2023attend}. 
Here, the user modification information, $\mathbf{F}_u$, serves as value and key, with the original feature map $\mathbf{F}_c$ acting as the query. 
\ourshort-v2 employs a transformer encoder to process $\mathbf{F}_u$, and its output, along with $\mathbf{F}_c$ is fed into a transformer decoder.  
The result is a dynamically updated feature map that effectively responds to significant pixel positions as per user interactions. This map, combined with $\mathbf{F}_c$, proceeds to a recalibration process akin to that in \ourshort-v1. 

\begin{figure}[t!]
\begin{center}
  \includegraphics[width=1.0\linewidth]{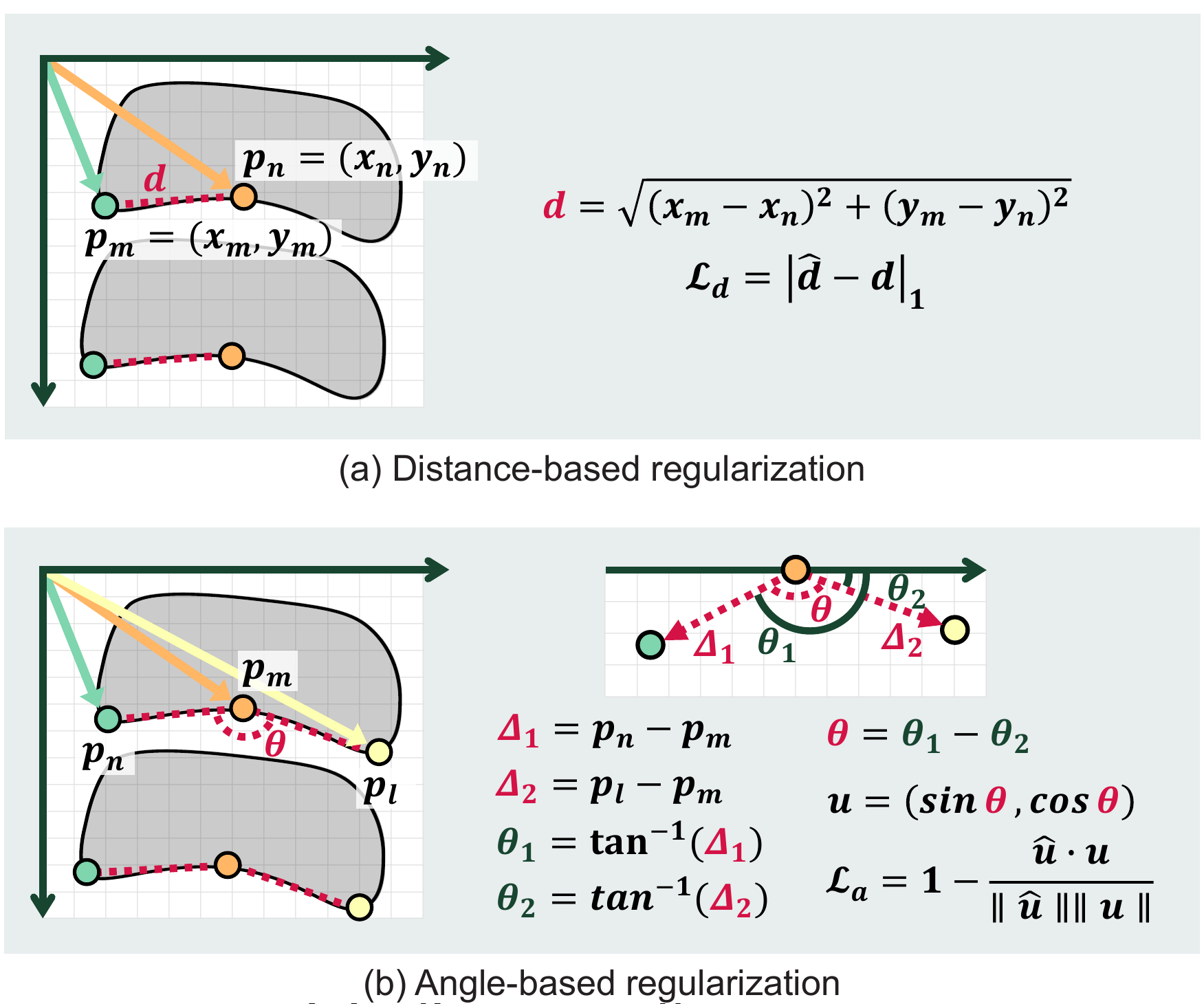}
\end{center}
\vspace{-10pt}
  \caption{\textbf{Morphology-aware loss.} A set of predefined keypoints, say, $p_n, p_m, p_l$, are used to regularize the model to preserve the consistent inter-keypoint relationships, focusing on (a) distance and (b) angle between keypoints.} 
\label{fig:morph_overview}
\end{figure}

\begin{figure*}[t!]
\begin{center}
  \includegraphics[width=1.0\linewidth]{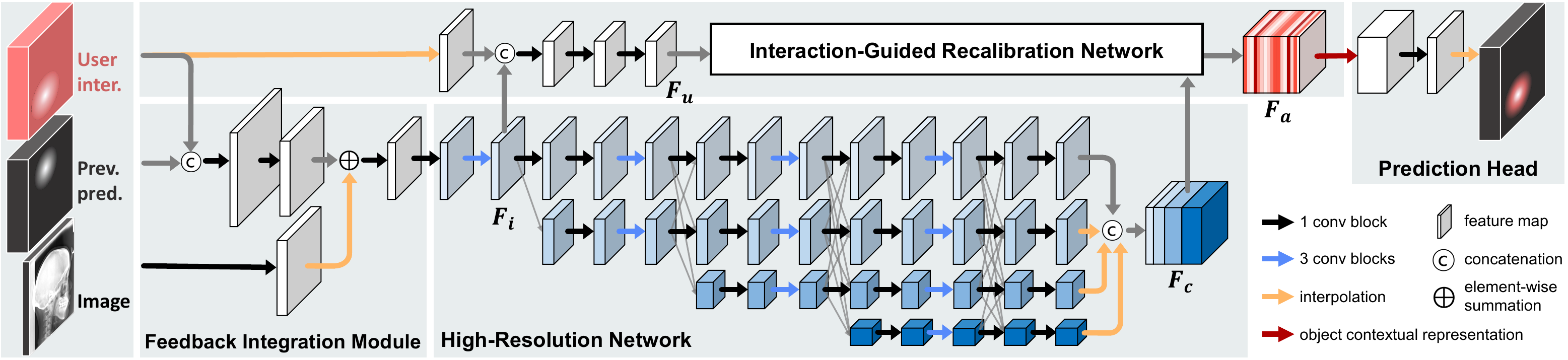}
\end{center}
\vspace{-10pt}
  \caption{\textbf{\ourmethod (\ourshort).} 
 Our model is designed to process radiographs alongside user interactions (\textit{User Inter.}) and its previous predictions (\textit{Prev. Pred.}). The model generates a heatmap of keypoint locations, which is dynamically adjusted to reflect user feedback.
}
\label{fig:main_network}
\end{figure*}
\subsection{Morphology-aware loss}\label{sec:morph}
Inspired by the observation that keypoints on the vertebrae exhibit limited deformation,
we introduce a morphology-aware loss function to maintain the morphological consistency of keypoints. 
This loss function enforces coherent updates across keypoints in response to user interactions, ensuring that their anatomical and spatial relationships are preserved in the refined model predictions, thereby enhancing the effectiveness of user interactions.

We represent inter-keypoint relationships using two higher-order statistics: the distance $d_{m,n}$ between two keypoints, $p_m$ and $p_n$, and the angle $u_{m,n,l}$ formed by three keypoints $p_m$, $p_n$, and $p_l$, as shown in Fig~\ref{fig:morph_overview}. 
For a total of $K$ keypoints, this results in $\binom{K}{2}$ distance combinations and $\binom{K}{3}$ angle combinations.

However, our method does not consider all possible combinations. We focus on subsets where the relationships exhibit minimal deviation across the dataset.
We select the subsets based on standard deviation values for distances and angles. Note that we employ the circular variance for angles.
Let $d$ and $u=(x_u, y_u)$ denote a distance and angle vector, respectively, as illustrated in Fig.~\ref{fig:morph_overview}. Then, the standard deviations are calculated as: 
\begin{equation}
\begin{aligned}
    S_d(d) &= \sqrt{\ \ \E\Big[\Big(d-\E[d]\Big)^2\Big]\ \ },  \\
    S_a(u) &= \sqrt{\ -\ln\Big(\E[x_u]^2+\E[y_u]^2\Big)\ }.
\end{aligned}
\end{equation}
The subsets $\mathcal{P}_d$ and $\mathcal{P}_a$ are then defined based on these standard deviations and threshold values $t_d$ and $t_a$ as: 
\begin{equation}
\begin{aligned}
        \mathcal{P}_d &= \Big\{\ \ \ (p_m, \ p_n)\  \ \  \big|\  \  S_d\Big(\,d_{m,n})\,\Big) < t_d, \\&\qquad \qquad\qquad \qquad
\ m \neq n; \ m, n \in {[1, K]} \ \ \ \Big\},\\ 
     \mathcal{P}_a &= \Big\{\ (p_m, p_n, p_l)\   \big|\   \ S_a\Big(\,u_{m,n,l})\,\Big) < t_a, \\&\qquad \qquad\qquad \qquad
    \ m \neq n \neq l; \  m, n, l \in [1,K]\ \Big\}. 
\end{aligned}
\end{equation}
For these subsets, we apply an $L1$ distance loss $L_{d}$ for $\mathcal{P}_d$, and a cosine similarity loss for angle vector $L_{a}$ for $\mathcal{P}_a$.
The overall morphology-aware loss is then defined as: 
\begin{equation}
   L_m = L_{d} + \lambda_a L_{a},  
\end{equation}
where $\lambda_a$ is a hyperparameter to set.
This design ensures that when a user corrects one keypoint, other keypoints with regular relationships to the corrected one are also updated, thereby preserving the morphological consistency within our model.

\subsection{Network architecture.}
\ourshort consists of three primary components: \attendOurs, feedback integration module, a backbone network, and a prediction head, as illustrated in Fig.~\ref{fig:main_network}.

Inspired by previous work~\citep{ritm}, the feedback integration module processes input radiographs, earlier model predictions, and user inputs. In the initial prediction, the model relies solely on the radiographs with user interactions and previous predictions set as zero. In the refinement phase, both of them are updated accordingly. Feeding the previous predictions as the model input enables the model to learn from its past outputs~\citep{mahadevanitis2018,ritm}. During inference, we enhance this method by selectively including predictions corresponding to user-modified keypoints. 
The backbone of our model is the pretained High-Resolution Network (HRNet)-W32~\citep{hrnetv2} augmented with Object-Contextual Representations (OCR)~\citep{ocr}. 
The prediction head decodes features from \attendOurs into precise heatmaps, from which keypoint coordinates are extracted. 
We apply post-processing to the prediction results to ensure that keypoints modified by users maintain their intended positions.

For training, \ourshort employs two loss functions: keypoint detection loss $L_{g}$ that measures alignment between the predicted and target heatmaps using a binary cross-entroy loss, and morphology aware loss $L_{m}$. These are combined as:
\begin{equation}
   L = L_g + \lambda_m L_m,
\end{equation}
where $\lambda_m$ is a pre-determined hyperparameter.

\subsection{Overall process}
\rbttwo{
Interactive keypoint estimation aims to refine inaccurate model predictions with minimal user input, allowing a single correction to revise multiple keypoints. 
% Our model is trained once and remains fixed during inference. 
It serves both initial prediction and subsequent error refinement in response to user corrections without updating its weights. }
\rbttwo{
Once trained, the inference pipeline follows three key steps:
}
\begin{itemize}
  \setlength\itemsep{-0.1em}
    \item \rbttwo{Initial keypoint estimation: Given (i) an X-ray image, the model generates (ii) an initial keypoint prediction.}
    \item \rbttwo{User Interaction: If necessary, (iii) a user corrects one of the predicted keypoints.}
    \item \rbttwo{Model revision: The model takes (i) the X-ray image, (ii) the initial prediction, and (iii) the user correction as inputs and refines the remaining errors accordingly (See Fig.~\ref{fig:main_network}).}
\end{itemize}

\rbttwo{
To enable interactive refinement without modifying model parameters at inference, the model is explicitly trained under simulated user correction scenarios. By leveraging simulated user interactions during training, the model learns how corrections should propagate, ensuring accurate and efficient keypoint estimation with minimal user input.
}

\rbttwo{
During training, the number of user interactions is sampled from the range $[0, K]$ to expose the model to diverse interaction scenarios with predefined probabilities. For example, the zero-hint scenario occurs with a $\frac{1}{8}$ probability, while the single-hint scenario is encountered with a $\frac{1}{2}$ probability. These scenarios can be categorized into two key groups:
}

\begin{itemize}
\setlength\itemsep{-0.1em}
    \item \rbttwo{Zero-hint scenario: The model learns to autonomously generate keypoint predictions based solely on the input image.}
    \item \rbttwo{Multiple-hint scenarios: The model learns to refine keypoints based on the input image, previous predictions, and simulated user corrections. The number of user-provided hints follows a decaying probability distribution to prioritize minimal user effort while ensuring exposure to a variety of interaction cases.}
\end{itemize}

\rbttwo{
After determining the number of hints, keypoints to be corrected are randomly selected, and their ground-truth coordinates are provided as simulated user inputs. The model then learns to refine all keypoints in response to user feedback without modifying its parameters.
}

\rbttwo{
By training under both zero-hint and multiple-hint scenarios, our model seamlessly integrates initial keypoint estimation and interactive refinement within a single, end-to-end learning framework. Once trained, the model remains unchanged duri ng inference, simply refining predictions based on learned correction patterns. This eliminates the need for costly re-training while still allowing for adaptive corrections.
}

\section{Growth potential estimation}\label{sec:growth}
Our research introduces a novel method for assessing a patient's remaining growth potential using a single lateral cephalometric radiograph. Our method provides diagnostic guidance in determining growth potential based on the morphology of the cervical vertebrae at the growth peak.

\begin{figure*}[t!]
\begin{center}
\includegraphics[width=1.0\linewidth]{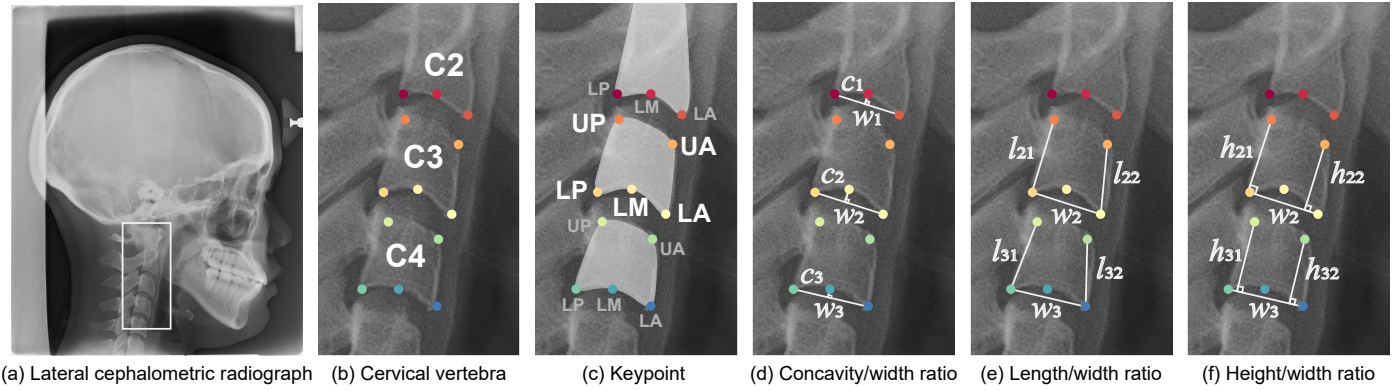}
\end{center}
\vspace{-10pt}
\caption{\textbf{Keypoints on the cervical vertebrae and \cvm features.} (a) Lateral cephalometric radiograph. (b) Thirteen keypoints on cervical vertebrae, upper posterior (UP), upper anterior (UA), lower posterior (LP), lower middle (LM), and lower anterior (LA). (c) Up to five vertex points per vertebra. (d-f) Measurements for \cvm estimation: concavity/width ratio (c/w ratio), length/width ratio (l/w ratio), and height/width ratio (h/w ratio).} 
\label{fig:keypoint}
\end{figure*}

\begin{figure*}[ht!]
\begin{center}
  \includegraphics[width=1.0\linewidth]{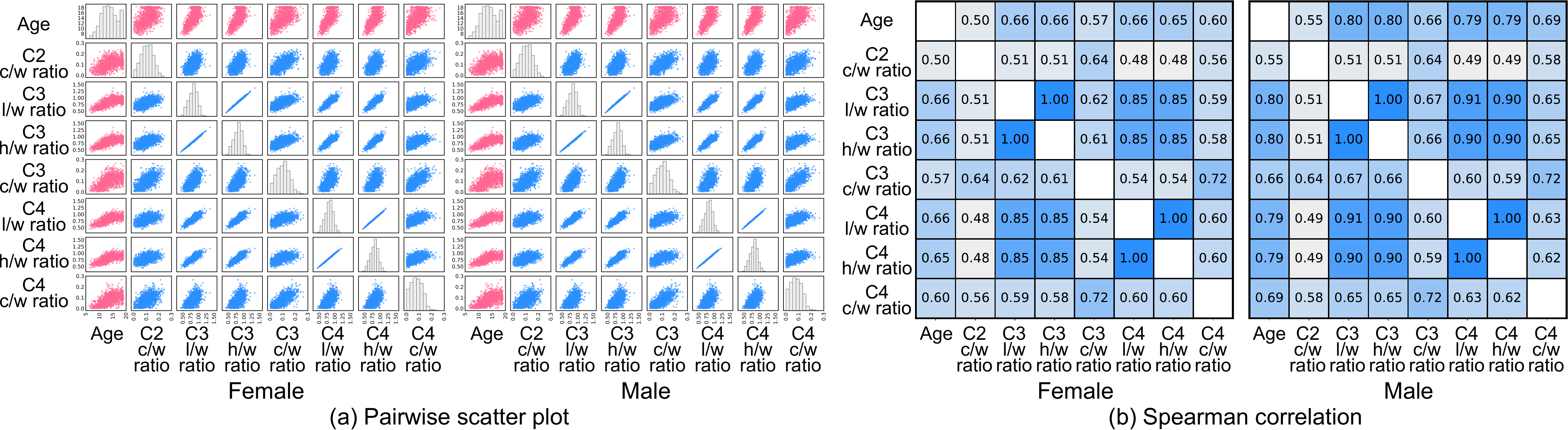}
\end{center}
\vspace{-10pt}
   \caption{\rbt{\textbf{Analysis of pairwise relationships between chronological age and \cvm features}: Concavity/width (c/w) ratio of C2, C3, and C4; length/width (l/w) ratio of C3 and C4; and height/width (h/w) ratio of C3 and C4. This analysis focuses on patients aged 6-18 years to examine the relationships during the growth phase. (a) Pairwise scatter plots visualize relationships between variables in the off-diagonal elements, while diagonal elements show individual variable distributions. (b) Spearman correlation matrices quantify the monotonic correlation between variables.}
}
\label{fig:corr}
\end{figure*}

\begin{figure*}[t!]
\begin{center}
  \includegraphics[width=1.0\linewidth]{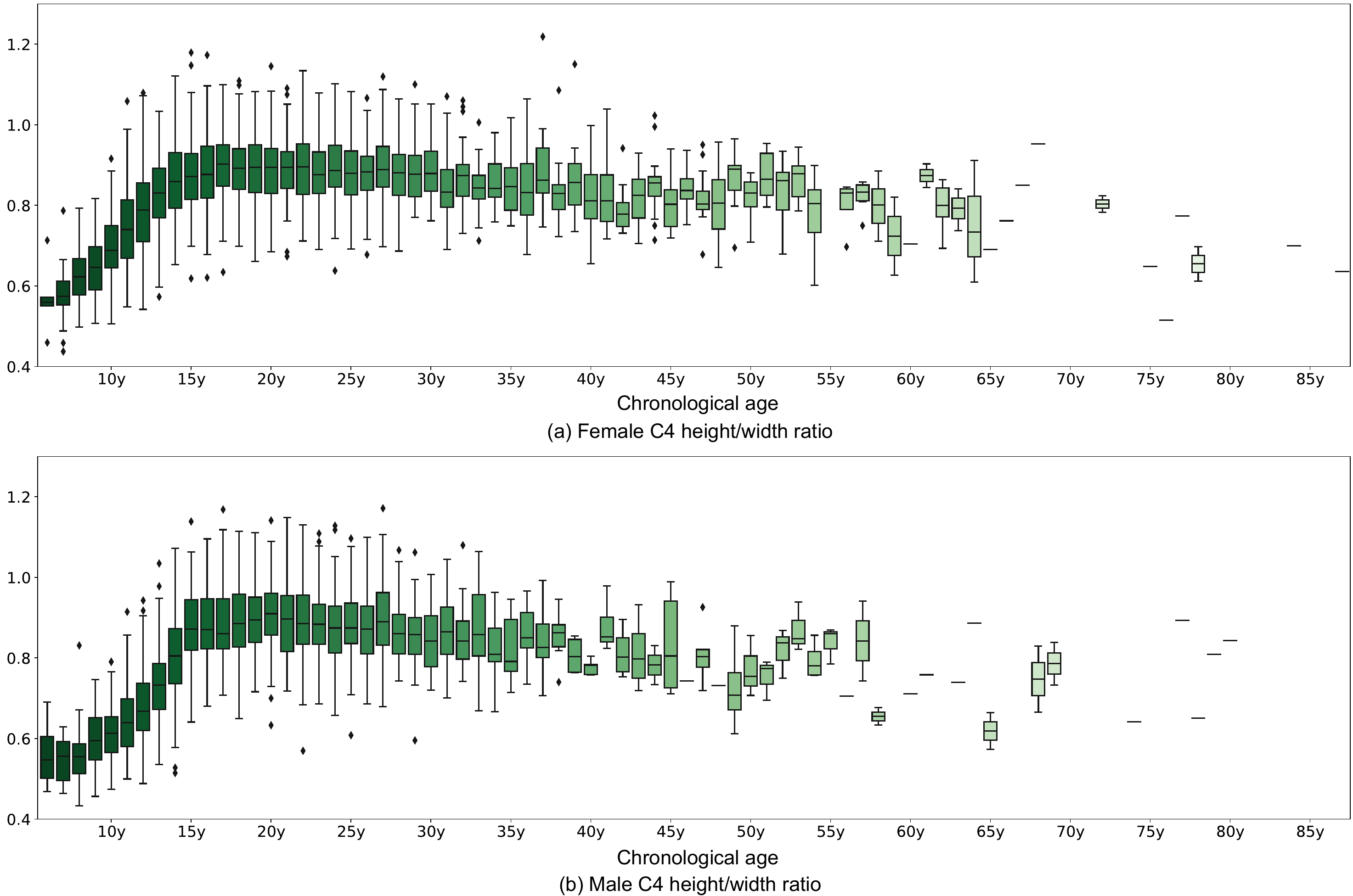}
\end{center}
\vspace{-10pt}
  \caption{\textbf{Standard growth curves for C4 height/width ratio.} Each box plot illustrates the interquartile range, marked by the first and third quartiles, with the median value indicated by a horizontal line inside the box. The whiskers extending from each box denote the minimum and maximum values.
}

\label{fig:growthcurve}
\end{figure*}

\subsection{Preliminary}\label{sec:preliminary}
This section describes the preliminary information about the keypoints on the cervical vertebrae and cervical vertebral maturation (CVM) feature values pivotal in assessing growth potential.

\subsubsection{Keypoints on the cervical vertebrae}
We focus on 13 keypoints located on the second (C2), third (C3), and fourth (C4) cervical vertebrae. These keypoints include the upper posterior (UP), the upper anterior (UA), the lower posterior (LP), the lower middle (LM), and the lower anterior (LA) points of each cervical vertebra, as shown in Figs.~\ref{fig:keypoint}a-c.

\subsubsection{Cervical vertebral maturation features.}\label{subsubsec:cvm}
The keypoints from cephalometric images enable us to analyze cervical vertebrae morphology and assess \cvm by estimating their morphological features.
Building upon the established  clinical diagnosis methods~\citep{cvm3,cvm2,kim2021prediction}, we use three morphometric features of cervical vertebrae for this analysis: concavity/width ratio (c/w ratio), length/width ratio (l/w ratio), and height/width ratio (h/w ratio). The measurements for these features, as shown in Fig.~\ref{fig:keypoint}d-f, are calculated for the $i$-th cervical vertebra can be obtained as: 
\begin{equation}
    \frac{c_i}{w_i}, \frac{l_{i1}+l_{i2}}{2w_i}, \text{ and }\frac{h_{i1}+h_{i2}}{2w_{i}},
\end{equation}
where $c_i$ represents the concavity, indicating the perpendicular distance from the most concave point to the lower border of the cervical vertebra body.

The length/width ratio and height/width ratio measure the aspect ratio of the cervical vertebra body, where $l_{i1}$ and $ l_{i2}$ represent the vertical distances from the upper to the lower point, and $h_{i1}$ and $h_{i2}$ denote the perpendicular distances from the upper point to the lower border. \rbt{The length/width ratio measures the average vertical length between the upper and lower vertices of the vertebrae, while the height/width ratio measures the average perpendicular height from the upper vertex to the base formed by the two lower vertices.}
In total, seven \cvm features are used for estimating growth potential.

\subsection{\rbt{Dataset for growth peak analysis}}\label{sec:datagrowth}
\rbt{For this study, we collect a dataset of 5,728 lateral cephalometric radiographs from two medical institutions. Since sex information is critical for accurate growth peak analysis, two images lacking this information are excluded from this specific analysis. Consequently, our growth peak analysis is conducted on a refined dataset comprising} 5,726 lateral radiographs from 3,812 individuals, as detailed in Table~\ref{table:data}. 
The dataset comprises $42.1\%$ male and $57.9\%$ female patients. Our analysis specifically targets the skeletal growth period of ages 6 to 18 years, which includes 2,679 cases in our dataset. Note that \textit{age} denotes chronological age in this paper, unless stated otherwise.
This substantial data volume allows our research to offer more robust and comprehensive insights, overcoming the limitations of previous studies that were often restricted by smaller sample sizes~\citep{lee2020structure,peng2020saint}.

Furthermore, we gathered a total of 499 pairs of hand-wrist radiographs and lateral cephalograms from two distinguished university hospitals and a specialized pediatric dental clinic. This dataset comprises images from 455 patients, including 208 males and 247 females. The patients' ages range from 6 to 18 years, with an average age of 9.9 years and a standard deviation of 2.6.

\subsection{Growth peak analysis}
\rbt{During the growth phase, the morphological characteristics of cervical vertebrae undergo notable changes, such as an increase in the length/width ratio. Analyzing changes in CVM features over age provides quantitative insights into these transformations.}
Our objective is to estimate the pubertal growth peak during childhood and adolescence by leveraging \cvm features, a critical assessment for determining the optimal timing for orthodontic interventions. 
Our methodology includes the analysis of standard growth curves and growth rates to identify the pubertal growth peak and investigate the morphological characteristics of cervical vertebrae at this pivotal stage. These traits can serve as reliable predictors of a patient's growth potential, offering valuable insights to guide clinical decision-making.  

Our analysis investigates the relationship between \cvm features and chronological age. We employ scatter plots and Spearman rank correlation~\citep{spearman1961proof}, as shown in Fig.~\ref{fig:corr}. 
The results indicate a positive linear correlation between CVM features and age. Additionally, \cvm features exhibit significant interrelatedness, underscoring their concurrent development with age. 
\rbt{For example, the length/width ratios and height/width ratios exhibit strong monotonic correlations, which is because both ratios measure the relative proportions of cervical vertebrae width and height. During growth, cervical vertebrae elongate vertically relative to their width. Thus, the length/width ratio and height/width ratio are effective for quantitatively capturing the morphological changes associated with vertebral growth. }

% data 표
\begin{table}[t!]
    \centering
    \caption{\textbf{Baseline characteristics of the enrolled patients used in the growth peak analysis study.} \textit{stdev} indicates standard deviation. }\label{table:data_table} \label{table:data}
    \resizebox{0.64\linewidth}{!}{%
    \begin{tabular}{ll}
        \toprule
        \multicolumn{1}{l}{\textbf{Total patient cohort}} & \textbf{Value} \\
        \midrule
        Patients, \textit{n} & 3,812 \\
        Images, \textit{n} & 5,726 \\
        Male, \textit{n} (\%) & 2,415 (42.1) \\
        Age, mean $\pm$ stdev (years) & 21.0 $\pm$ 9.8 \\
        \midrule
        \multicolumn{1}{l}{\textbf{Patients aged 6-18 years}} & \textbf{Value} \\
        \midrule
        % \multicolumn{2}{c}{\textbf{Patients aged 6-18 years}}\\
        Patients, \textit{n} &  1,864 \\
        Images, \textit{n} & 2,679 \\
        Male, \textit{n} (\%) & 1,227 (45.8) \\
        Age, mean $\pm$ SD (years) & 13.8 $\pm$ 2.9 \\
        % \multicolumn{1}{l}        {\textbf{Cervical vertebral maturation, mean $\pm$ SD}} \\
        % C2 concavity/width ratio & $\pm$ \\
        % C3 concavity/width ratio & $\pm$ \\
        % C3 length/width ratio & $\pm$ \\
        % C3 height/width ratio & $\pm$ \\
        % C4 concavity/width ratio & $\pm$ \\
        % C4 length/width & $\pm$ \\
        \bottomrule
    \end{tabular}
    }
\end{table}

\rbt{Although these metrics are correlated due to their shared focus, each provides unique insights into vertebral morphology. The key distinction lies in how height is measured (refer to Section~\ref{subsubsec:cvm}).
By analyzing both ratios, we capture complementary aspects of cervical vertebra morphology, offering a more comprehensive understanding of vertebral growth patterns. This dual perspective accounts for both vertical span and perpendicular height, leveraging the complementary strengths of these features. This ensures that no valuable information is excluded, resulting in robust and reliable analyses.
Therefore, while the strong correlations reflect a shared focus, they do not imply redundancy. Instead, including both ratios contributes to a thorough and robust growth peak analysis.}

Further corroborating these findings, we utilize box plots to represent the standard growth curves of CVM feature values using our extensive datasets, as shown in Fig.~\ref{fig:growthcurve}. 
These plots effectively illustrate the typical progression of CVM features with age, delineating the median, interquartile range, and extreme values. 
Significantly, \cvm features generally increase with age during the growth period, followed by a plateau, indicating an end of growth, followed by a decrease, indicating degenerative changes. 
In particular, males exhibit a more rapid growth rate in CVM features after age 10, consistent with their higher correlation values with age than females.
By comparing patients' CVM feature values against the standard growth curve for their age group, we can evaluate whether their growth in CVM is faster, slower, or in line with their peers. 
However, relying solely on the standard growth curves of CVM values for predicting growth potential is insufficient due to individual variations in growth trajectories. The ultimate CVM value at the end of the growth phase differs among individuals, meaning the CVM value alone is not a definitive indicator of growth potential.

Therefore, identifying the peak growth velocity (i. e. peak annual growth rate) becomes critical for accurately assessing the patient's skeletal growth stage and potential for growth. 
For example, a patient whose CVM values are in the first quartile for the same age group  (i.e. higher than the peers) may not have much residual growth if the patient's pubertal growth peak has passed. Conversely, a patient with lower CVM values compared to the same age group may be expected to have greater potential for growth if that person's growth rate has not passed the pubertal peak.
The growth peak is determined by analyzing annual growth rates or the slopes of the growth curves.
\rbt{In this study, we analyze growth peak timing using median values for each age group. The median is particularly well-suited for this analysis as it reflects general growth patterns within the population while minimizing the influence of outliers with extreme growth patterns that could skew results. This approach provides robust insights into population-level growth trends in cervical vertebrae. 
}

Upon identifying the age at the growth peak, we examine the morphological traits of cervical vertebrae at this peak to identify the specific cervical morphology indicative of this developmental stage.
This analysis allows us to make precise predictions about a patient's growth peak based on their current cervical vertebral morphology, as indicated by their CVM feature values. Consequently, clinicians can use this information to determine the optimal timing for orthodontic interventions.

Additionally, we aim to determine the primary CVM feature that can serve as a key diagnostic reference for skeletal maturation stages of the pubertal peak. To accomplish this, we analyze the correlation between CVM features and Fishman's skeletal maturation index (SMI) values obtained from hand-wrist radiographs. SMI has been used as a strong indicator of jaw growth. We specifically focus on cases with the SMI stage 4, which indicates the occurrence of the growth peak. This enables us to examine the CVM feature values at the time of the growth peak with increased reliability.
Our analysis involves 42 pairs of lateral cephalograms and hand-wrist radio-graphs with a SMI of 4 sourced from our dataset.  
This investigation allows us to determine which CVM feature should be used as the most suitable reference for diagnosis.

In conclusion, our approach, employing standard growth curves and growth rate analysis from an extensive dataset, estimates growth peak and analyzes cervical vertebrae morphology at these pivotal points. This approach facilitates accurate predictions of a patient's growth potential using a single lateral cephalometric radiograph, informed by their current cervical vertebra morphology.
Section~\ref{sec:results_growth} thoroughly discusses the results and detailed analysis.

\section{Experiments}
 This section describes the experimental setup and provides the quantitative and qualitative analysis of our proposed method.

\subsection{Experimental setup}
This section explains the dataset, evaluation metrics, baseline models, and implementation details. 

\subsubsection{\rbt{Dataset for interactive keypoint estimation}}\label{sec:dataset}
\rbt{To evaluate the effectiveness of the proposed interactive keypoint estimation model, we use our dataset of \textbf{\ourdata} (CEP), which is also utilized for growth peak analysis, as described in Section~\ref{sec:datagrowth}. Note that the entire dataset is used here, as sex information is not required for keypoint model development and evaluation. }
The dataset consists of 5,728 radiographs, randomly divided into training (546 images), validation (561 images), and test sets (4,621 images), with no overlap of patients across these sets. Each radiograph is annotated with 13 keypoints on the cervical vertebrae, as illustrated in Fig.~\ref{fig:keypoint}. These annotations are verified and confirmed by skilled dental residents and board-certified orthodontists.
 
To demonstrate the versatility and general applicability of our approach, we also validate our method on public datasets. 
The \textbf{\spinewebdata} dataset~\citep{spinewebdataset} contains spinal anterior-posterior radiographs consisting of 352 training images, 128 validation images, and 128 test images. Each image is annotated with 68 points representing the four vertices of 17 vertebrae. 

The BUU SPINE dataset~\citep{klinwichit2023buu} consists of a pair of images with different views collected from 400 unique patients. 
\textbf{\buudataAP} contains 400 radiographs captured in the anterior-posterior (AP) view. Each image in this set is meticulously annotated with 20 keypoints, identifying crucial anatomical landmarks on the edges of the lumbar spine. We split the data into 240 images for training, 80 for validation, and 80 for testing. The image sizes range from (1434, 1072) to (3072, 3040). 
\textbf{\buudataLA} includes 400 radiographs taken from the left lateral (LA) view. Each image is annotated with 22 spinal keypoints. Similar to \buudataAP, we split the dataset into 240, 80, and 80 images for the training, validation, and test sets, respectively. The image sizes in this dataset range from (1956, 968) to (3072, 3040).

\begin{table*}[t!]
\caption{\textbf{Performance comparison of interactive keypoint estimation on the AASCE, BUU-AP, and BUU-LA datasets.}}
\vspace{-10pt}
% compared with state-of-the-art methods.} }
%$\dag$ denotes that the model performance without a backpropagation refinement scheme is reported.}
\label{table:ikepublic}
\begin{center}
\resizebox{1.0\textwidth}{!}{%
\begin{tabular}{l|ccccc|ccccc|ccccc}
\toprule
% ===== 맨위 ===========
\multicolumn{1}{l}{\multirow{3.5}{*}{Method}} & 
\multicolumn{5}{c}{\spinewebdata} &
\multicolumn{5}{c}{\buudataAP} &
\multicolumn{5}{c}{\buudataLA} \\
\cmidrule(l{2pt}r{2pt}){2-6} \cmidrule(l{2pt}r{2pt}){7-11}\cmidrule(l{2pt}r{2pt}){12-16}
\multicolumn{1}{c}{}&
\makecell{$\text{FR}_{10}$\\@20} & 
\makecell{$\text{NoC}_{10}$\\@20} &
\makecell{$\text{NoC}_{10}$\\@30} & 
\makecell{$\text{NoC}_{10}$\\@40} &
\makecell{$\text{NoC}_{10}$\\@50} &
\makecell{$\text{FR}_5$\\@6} & 
\makecell{$\text{NoC}_5$\\@6} &
\makecell{$\text{NoC}_5$\\@8} & 
\makecell{$\text{NoC}_5$\\@10} &
\makecell{$\text{NoC}_5$\\@12} &
\makecell{$\text{FR}_5$\\@6} & 
\makecell{$\text{NoC}_5$\\@6} &
\makecell{$\text{NoC}_5$\\@8} & 
\makecell{$\text{NoC}_5$\\@10} &
\makecell{$\text{NoC}_5$\\@12} \\ 
% ================== 이제 성능 기록 ========
\midrule\midrule
\multicolumn{1}{l|}{$\text{BRS}$} & \blue{11.72} & \blue{3.02} & \blue{2.41} & \blue{1.91} & \blue{1.59}
& \blue{100.00} & \blue{5.00} & \blue{4.83} & \blue{4.54} & \blue{3.88} %ts412
& \blue{100.00} & \blue{5.00} & \blue{4.97} & \blue{4.79} & \blue{4.58} %ts411
\\
\multicolumn{1}{l|}{$\text{f-BRS}$} & \blue{52.34} & \blue{7.36} & \blue{5.61} & \blue{4.55} & \blue{3.82}
& \blue{96.25} & \blue{4.95} & \blue{4.71} & \blue{4.19} & \blue{3.27} %ts418
& \blue{81.25}& \blue{4.70}& \blue{4.01}& \blue{3.02}& \blue{2.12}%ts417
\\
\multicolumn{1}{l|}{RITM} & \blue{13.28} & \blue{3.55} & \blue{2.73} & \blue{2.14} & \blue{1.56}
& \blue{78.75} & \blue{4.67} & \blue{3.67} & \blue{2.58}  &\blue{1.86} %ts415
& \blue{71.25}& \blue{4.45}& \blue{3.21}& \blue{1.98}& \blue{1.10} %ts416
\\
\rbt{ClickPose} & \rbt{57.03} & \rbt{6.51} & \rbt{5.21} & \rbt{4.59} & \rbt{4.09} & {\rbt{58.75}} & \rbt{4.34} & \rbt{3.30} & \rbt{2.31} & \rbt{1.68} & \rbt{57.50} & \rbt{4.25} & \rbt{3.11} & \rbt{1.91} & \rbt{1.24}\\
% \rbt{Dai et al.} & \rbt{100.0} & \rbt{10.00} & \rbt{10.00} & \rbt{10.00} & \rbt{9.85} & \rbt{56.25} & \rbt{4.24} & \rbt{3.36} & \rbt{2.73} & \rbt{2.40} & \rbt{100.0} & \rbt{5.00} & \rbt{5.00} & \rbt{5.00} & \rbt{5.00}\\
\midrule  
%&
% HRNet-W18 & 10.20 & 3.205 & 1.536 & 0.630 & 0.262 &  \textbf{5.47} & 2.633 & \textbf{1.750} & \textbf{1.227} & \textbf{0.977}\\
\multicolumn{1}{l|}{{\ourshort-v1}}&  \blue{6.25} & \blue{2.46} & \blue{1.88} & \blue{1.41} & \blue{1.19}
& \blue{65.00} & \blue{4.45} & \blue{3.44} & \blue{2.40} & \blue{1.65} % 398
& \blue{61.25} & \blue{4.31} & \blue{3.01} & \blue{1.70} & \blue{\textbf{0.88}} % 407
\\
% \midrule  
\multicolumn{1}{l|}{{\ourshort-v2}}&\blue{\textbf{3.91}} & \blue{\textbf{1.94}} & \blue{\textbf{1.44}} & \blue{\textbf{1.10}} & \blue{\textbf{0.89}}%225 
&\blue{\textbf{57.50}}&\blue{\textbf{4.28}}&\blue{\textbf{3.20}}&\blue{\textbf{2.01}}&\blue{\textbf{1.31}} %495
& \blue{\textbf{56.25}} &\blue{\textbf{4.24}}&\blue{\textbf{2.76}}&\blue{\textbf{1.61}}&\blue{\textbf{0.88}} %394 %V2-3
\\ 
% \multicolumn{1}{l|}{{\ourshort-v2-2}}& {6.25} & {1.88} & {1.41} & {1.16} & {0.94}%227
% & running& & & & % 
% & running &&&& %
% \\ 
% \multicolumn{1}{l|}{{\ourshort-v2-1}}&{4.69} & \textbf{1.57} & \textbf{1.22} & {1.06} & {0.89} %224
% & running& & & & % 
% & running &&&& %
% \\
\bottomrule
  \end{tabular}}
\end{center}
\end{table*}

\begin{table}[t!]
\caption{\textbf{Performance comparison of interactive keypoint estimation on the \ourdata dataset.}}
\vspace{-10pt}
\label{table:ikeours}
\begin{center}
\resizebox{1.0\linewidth}{!}{%
\begin{tabular}{l|ccccc}
\toprule
% ===== 맨위 ===========
\multicolumn{1}{l}{\multirow{2.5}{*}{Method}} & 
\multicolumn{5}{c}{CEP}  \\
\cmidrule(l{2pt}r{2pt}){2-6} 
\multicolumn{1}{c}{}&
% \makecell{$\text{FR}_5$\\@3} & 
% \makecell{$\text{NoC}_5$\\@3} &
% \makecell{$\text{NoC}_5$\\@4} & 
% \makecell{$\text{NoC}_5$\\@5} &
% \makecell{$\text{NoC}_5$\\@6}  \\ 
\makecell{$\text{FR}_5$@3} & 
\makecell{$\text{NoC}_5$@3} &
\makecell{$\text{NoC}_5$@4} & 
\makecell{$\text{NoC}_5$@5} &
\makecell{$\text{NoC}_5$@6}  \\ 
% ================== 이제 성능 기록 ========
\midrule\midrule
\multicolumn{1}{l|}{$\text{BRS}$} &24.41&3.69&2.05&0.89&0.37\\%ts425
\multicolumn{1}{l|}{$\text{f-BRS}$} & 5.95 & 2.61 & 1.03 & 0.37 & 0.14\\%ts422
\multicolumn{1}{l|}{RITM} & 12.03 & 3.21 & 1.53 & 0.63 & 0.25 \\%ts423
\multicolumn{1}{l|}{\rbt{Click-Pose}} & \rbt{33.00} & \rbt{{2.89}} & \rbt{2.25} & \rbt{1.80} & \rbt{1.47} \\
\midrule  
%&
% HRNet-W18 & 10.20 & 3.205 & 1.536 & 0.630 & 0.262 &  \textbf{5.47} & 2.633 & \textbf{1.750} & \textbf{1.227} & \textbf{0.977}\\
% \multicolumn{1}{l|}{\ourshort-v1}& {4.48} & {2.32} & {0.86} & {0.31} & {0.13}\\
\multicolumn{1}{l|}{\ourshort-v1}&  {5.17} & {2.56} & {1.00} & {0.35} & {0.13}  \\ %256
% \midrule  
% \multicolumn{1}{l|}{{\ourshort-v2-3}}& \textbf{4.74} & \textbf{2.36} & \textbf{0.89} & \textbf{0.31} & \textbf{0.12}\\ %265
% \multicolumn{1}{l|}{{\ourshort-v2-3(2)}}
% & 5.11 & 2.43 & 0.93 & 0.32 & 0.12 %357
% \\
\multicolumn{1}{l|}{\ourshort-v2}
& \textbf{4.54} & \textbf{2.37} & \textbf{0.90} & \textbf{0.31} & \textbf{0.12} %366 c2
\\
% \multicolumn{1}{l|}{\ourshort-v2-1}
% & 5.02 & 2.47 &0.95 &0.33&0.12 %294
% \\
% \multicolumn{1}{l|}{\ourshort-v2-3 48}
% & 4.74&2.42&0.91 &0.30&0.11 %372
% \\
% \multicolumn{1}{l|}{251 cep_combine}
% % % \multicolumn{1}{l|}{368 combine3}
% & 5.00 & 2.50 & 0.97 & 0.34 & 0.13 %368
% \\
\bottomrule
\end{tabular}}
\end{center}
\end{table}

\subsubsection{Evaluation metrics}
Keypoint estimation performance is measured by the mean radial error (MRE), following the previous work~\citep{ISBI}. 
Denoting the predicted and groundtruth keypoints as $p_n$ and $\hat{p}_n$, respectively, MRE is calculated as:
\begin{equation}
 \text{MRE}=\frac{1}{K}\sum_{n=1}^K(||p_n-\hat{p}_n||_2).
\end{equation}

To assess the effectiveness of our interactive keypoint estimation approach, we adopt the evaluation protocol used in interactive segmentation tasks~\citep{firstclick,fbrs}: the number of clicks (NoC) and the failure rate (FR).
NoC evaluates the average number of user clicks required to achieve a target MRE. Specifically, $\text{NoC}_{\alpha}@\beta$ represents the average number of clicks needed to reach a target MRE of $\beta$, given that a prediction can be modified by up to $\alpha$ clicks. If a model fails to achieve the target MRE $\beta$ even with $\alpha$ clicks, we set the NoC value as $\alpha$. 
Our study aims to minimize the number of user interventions; therefore, the maximum number of user modifications is capped at a relatively low value. For datasets [\ourdata, \spinewebdata, \buudataAP, \buudataLA] with [13, 68, 20, 22] target keypoints, the limits are set to [5, 10, 5, 5] clicks, respectively.
Similarly, $\text{FR}_{\alpha}@\beta$ measures the rate of images that do not achieve the target MRE of $\beta$ pixels within the maximum allowed $\alpha$ clicks. 
During an evaluation, the keypoint with the highest error in each prediction is chosen for revision, assuming that a clinician would prioritize correcting the most evidently erroneous keypoints.

\begin{table*}[t!]
\caption{\rbt{
\textbf{Performance comparison of mean radial error across four datasets.} We compare manual revision (\texttt{Manual}) and model revision (\texttt{Model}). For Dai et al., model revision results are not provided, as no interactive module was proposed.}}\label{table:hint0hint1}
\small
\vspace{-10pt}
\begin{center}%\resizebox{1.0\textwidth}{!}{%
\begin{tabular}{l|cc|cc|cc|cc}
\toprule
% \multirow{2.3}{*}{Method} &
% \multicolumn{1}{c}{\multirow{3.5}{*}{\makecell{Selective \\ previous\\ prediction}}} &
% \cmidrule(lr ){13-17}
% \multicolumn{1}{c}{} & 
\multicolumn{1}{l}{\multirow{2.5}{*}{\rbt{Method}}}
&\multicolumn{2}{c}{\rbt{AASCE}}&\multicolumn{2}{c}{\rbt{BUU-AP}}&\multicolumn{2}{c}{\rbt{BUU-LA}}&\multicolumn{2}{c}{\rbt{CEP}}\\
\cmidrule(l{2pt}r{2pt}){2-3}
\cmidrule(l{2pt}r{2pt}){4-5}
\cmidrule(l{2pt}r{2pt}){6-7}
\cmidrule(l{2pt}r{2pt}){8-9}
\multicolumn{1}{c}{}&\texttt{\rbt{Manual}}&\texttt{\rbt{Model}}
&\texttt{\rbt{Manual}}&\texttt{\rbt{Model}}
&\texttt{\rbt{Manual}}&\texttt{\rbt{Model}}
&\texttt{\rbt{Manual}}&\texttt{\rbt{Model}}\\
\midrule \midrule
\rbt{BRS} & \rbt{49.90} & \rbt{42.42} & \rbt{44.46}  & \rbt{31.19} & \rbt{35.06}  & \rbt{43.19} & \rbt{0.604}  & \rbt{0.729}\\
\rbt{f-BRS} & \rbt{64.52}  & \rbt{63.77} & \rbt{38.65}  & \rbt{25.54} & \rbt{32.15} & \rbt{27.93} & \rbt{0.521}  & \rbt{0.629}\\
\rbt{RITM} & \rbt{59.78}  & \rbt{43.33} & \rbt{32.55}  & \rbt{26.06} & \rbt{20.55}  & \rbt{17.79} & \rbt{0.587}  & \rbt{0.583}\\
\rbt{ClickPose} & \rbt{62.64}  & \rbt{62.63} & \rbt{\textbf{29.57}}  & \rbt{29.57} & \rbt{22.44}  & \rbt{22.44} & \rbt{6.227}  & \rbt{6.200}\\
\rbt{Dai et al.} & \rbt{66.45}   & \rbt{-} & \rbt{47.00}  & \rbt{-} & \rbt{47.30}  & \rbt{-} & \rbt{0.896}  & \rbt{-}\\
\midrule
\rbt{ARNet-v1} & \rbt{55.85}  & \rbt{35.39} & \rbt{38.63} & \rbt{23.43} & \rbt{20.81} & \rbt{14.29} & \rbt{0.522}  & \rbt{0.516}\\
\rbt{ARNet-v2} & \rbt{\textbf{46.04}}  & \rbt{\textbf{32.02}} & \rbt{30.47} & \rbt{\textbf{16.93}} & \rbt{\textbf{17.30}}  & \rbt{\textbf{13.45}} & \rbt{\textbf{0.515}}  & \rbt{\textbf{0.507}}\\
\bottomrule
\end{tabular}%}
\end{center}
\end{table*}

\begin{figure*}[t!]
\begin{center}
\includegraphics[width=1.0\linewidth]{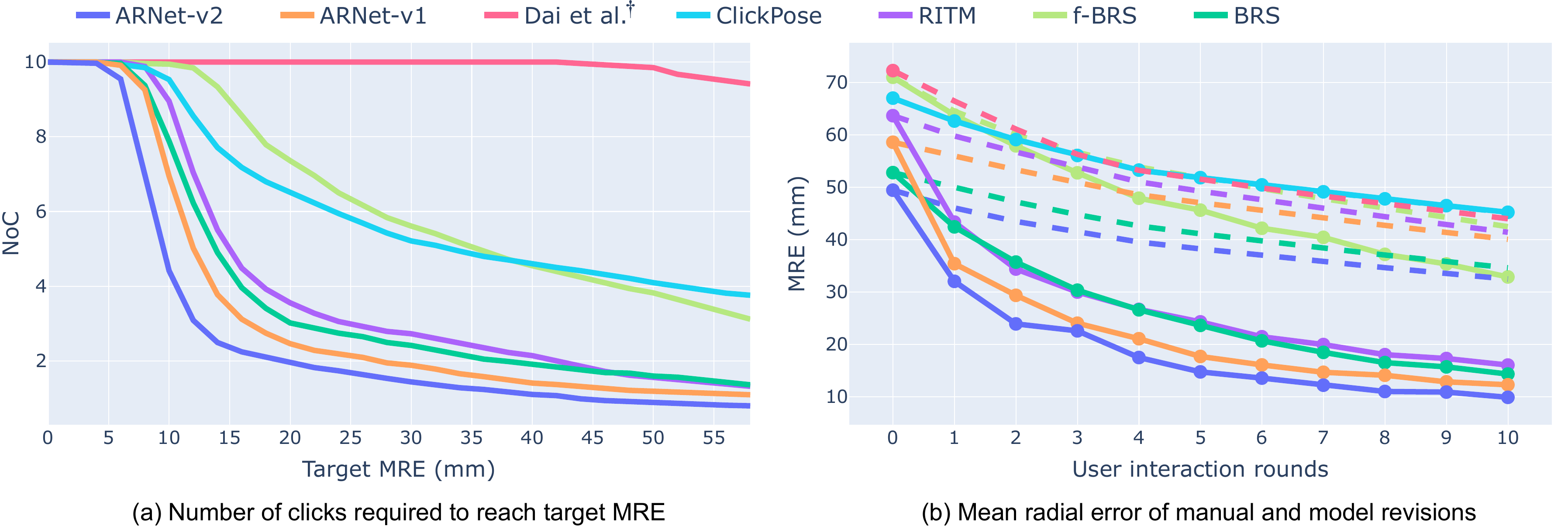}
\end{center}
\vspace{-10pt}
\caption{\rbt{\textbf{Comparison with state-of-the-art models on the AASCE dataset.} (a) Keypoint prediction errors for increasing number of user interactions in comparison with existing baseline models. (b) Number of user interactions for increasing target MRE in comparison with existing baseline models.}
}
\label{fig:noc}
\end{figure*}

\subsubsection{Compared baseline models.}
\noindent
\textbf{Comparison with interaction models.}
\rbt{\rbtthree{We evaluate ARNet-v2 against the state-of-the-art (SOTA) deep learning-based interactive models. ARNet-v1~\citep{kim2022morphology}, our previous work, is the first interactive keypoint model that achieves SOTA performance on cervical vertebrae keypoint datasets.
ClickPose~\citep{yang2023neural_clickpose} is an interactive keypoint estimation method designed for human pose estimation, aimed at refining multiple keypoints with minimal corrections. 
Dai et al.~\citep{dai2025cephalometric_daietal} proposes a dual-encoder-based keypoint regression method for high-resolution radiograph images, enhancing cephalometric keypoint estimation performance. Note that, since the model proposed by Dai et al. does not support interactive corrections, it has been excluded from the model interaction performance comparisons presented in Tables~\ref{table:ikepublic},~\ref{table:ikeours}, and~\ref{table:hint0hint1}.}}

\rbt{\rbtthree{Additionally, given the lack of interactive keypoint estimation models, our evaluation includes top-performing interactive segmentation approaches, such as BRS~\citep{brs}, f-BRS~\citep{fbrs}, and RITM~\citep{ritm}, with RITM particularly noted for its robust performance~\citep{kirillov2023segment_SAM_segment_anything}. 
  These approaches share similar concepts with our proposed interactive keypoint estimation framework in the context of interactive systems.   
  To adapt the interactive segmentation models to our interactive keypoint estimation task, we apply modifications that include adjustments to the user interaction representation, output layers, and loss function (refer to Appendix~\ref{appen_sec:baseline_reproducibility} for more details).}}

\noindent
\textbf{Manual Revision vs. Model Revision.} 
A crucial aspect of our evaluation involves comparing manual revision with model-assisted refinement in our proposed interactive keypoint estimation framework. Manual revision represents the traditional approach, where users adjust keypoints without the assistance of an interactive model, while model revision refers to addressing remaining errors following manual revision. This comparison highlights the improvements in efficiency and accuracy introduced by our proposed framework to the keypoint estimation process.

\subsubsection{Implementation details.}
We resize the images to different dimensions based on the dataset: $(512 \times 512)$ for \ourdata, and $(512 \times 256)$ for \spinewebdata, \buudataAP, and \buudataLA, respectively. These dimensions are chosen to optimize the balance between image detail and computational efficiency.
All experiments are conducted on a single GeForce RTX 3090. We utilize Python 3.8 and PyTorch 1.10.1 for the training and evaluation phases. 
For model optimization, we employ the AdamW optimizer~\citep{loshchilovdecoupled} with a learning rate of 0.001 and a batch size of 4. 

\subsection{Ethical approval}
The study protocol, \rbt{including data collection from patients (lateral cephalograms, hand-wrist radiographs, and demographic information)}, was granted approval by the institutional review boards (IRBs) of Korea University Anam Hospital (IRB no. 2019AN0394) and Korea University Guro Hospital (IRB no. 2020GR0071).

\subsection{Experimental results and analysis}
This section presents comprehensive experimental results on interactive keypoint estimation and growth peak analysis.

\subsubsection{Interactive keypoint estimation performance}  
\noindent
\textbf{State-of-the-art interaction performance.}
We evaluate the interactive keypoint estimation performance of \ourshort-v2 against baseline models by assessing their ability to improve annotation efficiency. As shown in Tables~\ref{table:ikepublic} and~\ref{table:ikeours}, ARNet-v2 achieves state-of-the-art results in failure rate (FR) and number of clicks (NoC) across four datasets, surpassing ARNet-v1 and baseline models. For example, on the AASCE dataset, ARNet-v2 reduces the failure rate by 37\% compared to ARNet-v1.
\ourshort-v2 consistently demonstrates a substantial reduction in failure rates while requiring significantly fewer user inputs to achieve the desired accuracy. On the AASCE dataset, ARNet-v2 shows a failure rate of just 3.91\%, reducing it by more than half compared to the baseline models.
\rbt{Additionally, ARNet-v2 consistently outperforms the only interactive keypoint estimation baseline, ClickPose, across all four datasets. For example, on the AASCE and CEP datasets, ClickPose exhibits high failure rates of 57.0\% and 33.0\%, respectively, while ARNet-v2 achieves significantly lower rates of 3.9\% and 4.5\%. }
These results highlight the broad applicability and remarkable superiority of our approach over existing methods, affirming its effectiveness in reducing failure rates and minimizing user interactions.

\begin{table*}[t!]
\footnotesize
\caption{\textbf{{Ablation study using \ourshort-v2 on the AASCE and BUU-AP datasets}.}}\label{table:ablation}
\vspace{-10pt}
\begin{center}\resizebox{1.0\textwidth}{!}{%
\begin{tabular}{ccc|ccccc|ccccc}
\toprule
% \multicolumn{1}{c}{\attendOurs} &
\multicolumn{3}{c}{Method} &
% \multicolumn{1}{c}{\multirow{3.5}{*}{\makecell{Selective \\ previous\\ prediction}}} &
\multicolumn{5}{c}{\spinewebdata}&
\multicolumn{5}{c}{\buudataAP}\\
\cmidrule(lr ){1-3}
\cmidrule(lr ){4-8}
\cmidrule(lr ){9-13}
% \cmidrule(lr ){13-17}
% \multicolumn{1}{c}{} & 
\multicolumn{1}{c}{\makecell{Cross-attention\\mechanism}} &
\multicolumn{1}{c}{\makecell{Gating\\mechanism}} & 
\multicolumn{1}{c}{\makecell{Morphology-\\aware loss}} &
\multicolumn{1}{c}{\makecell{$\text{FR}_{10}$\\$@20$}}&
\multicolumn{1}{c}{\makecell{$\text{NoC}_{10}$\\$@20$}}&
\multicolumn{1}{c}{\makecell{$\text{NoC}_{10}$\\$@30$}}&
\multicolumn{1}{c}{\makecell{$\text{NoC}_{10}$\\$@40$}}&
\multicolumn{1}{c}{\makecell{$\text{NoC}_{10}$\\$@50$}}&
\multicolumn{1}{c}{\makecell{$\text{FR}_{5}$\\$@6$}}&
\multicolumn{1}{c}{\makecell{$\text{NoC}_{5}$\\$@6$}}&
\multicolumn{1}{c}{\makecell{$\text{NoC}_{5}$\\$@8$}}&
\multicolumn{1}{c}{\makecell{$\text{NoC}_{5}$\\$@10$}}&
\multicolumn{1}{c}{\makecell{$\text{NoC}_{5}$\\$@12$}}\\
\midrule \midrule
\cmark & \cmark & \cmark 
&\blue{\textbf{3.91}}&\blue{\textbf{1.94}}&\blue{\textbf{1.44}}&\blue{\textbf{1.10}}&\blue{\textbf{0.89}} %225 c3
&\blue{\textbf{57.50}}&\blue{\textbf{4.28}}&\blue{\textbf{3.20}}&\blue{\textbf{2.01}}&\blue{\textbf{1.31}} \\ %495
\xmark& {\cmark}& {\cmark}    
&\blue{6.25}&\blue{2.46}&\blue{1.88}&\blue{1.41}&\blue{1.19} %ours 
&\blue{65.00}&\blue{4.45}&\blue{3.44}&\blue{2.40}&\blue{1.65}\\ %398
\cmark &   \xmark & \cmark  
&\blue{\textbf{3.91}}&\blue{2.08}&\blue{1.57}&\blue{1.34}&\blue{1.16} %457 no gating
&\blue{60.00}&\blue{4.39}&\blue{3.29}&\blue{2.26}&\blue{1.59} \\ %437  
\cmark & \cmark&  \xmark    
&\blue{7.81}&\blue{2.12}&\blue{1.58}&\blue{1.20}&\blue{0.99} %434 nomorph
&\blue{60.00}&\blue{4.47}&\blue{3.39}&\blue{2.35}&\blue{1.56} \\ %487
\xmark &  \xmark & \xmark  
&\blue{13.28}&\blue{3.55}&\blue{2.73}&\blue{2.14}&\blue{1.56} %RITM
&\blue{78.75}&\blue{4.67}&\blue{3.67}&\blue{2.58}&\blue{1.86} \\ % RITM32 (ts415)
% \cmidrule(l{1.5pt}r{1.5pt}){2-9}
% & max &sigmoid  &\cmark &low stdev. & - & \\
% \midrule
% \multirow{1}{*}{U-Net}
% & \cmark &  max & sigmoid  &\cmark&low variance & \\
\bottomrule
%%%%%%%%%%%%%%% Morph pairs
% & max &sigmoid& 5&5 &&\\
% & max &sigmoid&20&20&&\\
% \cmidrule(l{1.5pt}r{1.5pt}){2-8}
% & max &sigmoid&neighbor& 10&10 & &\\
%
\end{tabular}}
\end{center}
\end{table*}

\noindent
\textbf{{Manual vs. Model revision performance.}}
\rbtthree{While interactive keypoint estimation has clear benefits, it is not always reliable. For example, as shown in Fig.~\ref{suppe_fig:qual}, BRS occasionally worsens the predictions after incorporating user input, leading to increased error. To evaluate the practical effectiveness of our approach,}
\rbt{we compare interaction models by evaluating their ability to utilize user interactions to reduce errors. Specifically, we analyze the mean radial error (MRE) after one manual revision and after an additional model-assisted revision. The difference between these two values indicates how effectively each interaction model leverages user interactions to address remaining errors. As shown in Table~\ref{table:hint0hint1}, ARNet-v2 achieves the lowest MRE after model revision, outperforming all baseline models across all four datasets. In contrast, other baseline models demonstrate minimal improvements from model revision compared to manual revision. 
Notably, ClickPose exhibits a limited capacity to reduce errors through model-assisted interaction. Specifically, ARNet-v2 reduces errors compared to ClickPose by 48.9\% (AASCE), 42.7\% (BUU-AP), 40.06\% (BUU-LA), and 91.8\% (CEP). These results demonstrate the superior ability of ARNet-v2 to effectively utilize user interaction information.  }

\begin{table*}[t!]
\footnotesize
\caption{\textbf{Sensitivity analysis of \attendOurs and Morpholgy-aware loss using \ourshort-v2 on the AASCE and BUU-AP datasets}.
We apply different criteria for $\mathcal{P}_d$ and $\mathcal{P}_a$. For \textit{low $t_d, t_a$}, morphology-aware loss targets keypoint sets with the lowest variance, while \textit{high $t_d, t_a$} targets those with the highest variance. The \textit{adjacency} criterion focuses on keypoint sets forming an edge or internal angle of each vertebra.}
\vspace{-10pt}
\begin{center}\resizebox{1.0\textwidth}{!}{%
\begin{tabular}{cccc|ccccc|ccccc}
\toprule
% \multicolumn{1}{c}{\attendOurs} &
\multicolumn{3}{c}{\attendOurs} &
\multicolumn{1}{c}{\multirow{1}{*}{\makecell{Morphology-aware loss}}}&
% \multicolumn{1}{c}{\multirow{3.5}{*}{\makecell{Selective \\ previous\\ prediction}}} &
\multicolumn{5}{c}{\spinewebdata}&
\multicolumn{5}{c}{\buudataAP}\\
\cmidrule(lr ){1-3}
\cmidrule(lr ){4-4}
\cmidrule(lr ){5-9}
\cmidrule(lr ){10-14}
% \multicolumn{1}{c}{} & 
% \multicolumn{1}{c}{\makecell{Cross-\\attention}} &
\multicolumn{1}{c}{\makecell{Down\\sample}} &
% \multicolumn{1}{c}{Gating} &
\multicolumn{1}{c}{\makecell{Pooling\\method}} & 
\multicolumn{1}{c}{\makecell{Activation\\function}} & 
% \multicolumn{1}{c}{\makecell{Morph\\loss}} &
\multicolumn{1}{c}{\makecell{Criterion for\\$\mathcal{P}_d$ and $\mathcal{P}_a$}} &
\multicolumn{1}{c}{\makecell{$\text{FR}_{10}$\\$@20$}}&
\multicolumn{1}{c}{\makecell{$\text{NoC}_{10}$\\$@20$}}&
\multicolumn{1}{c}{\makecell{$\text{NoC}_{10}$\\$@30$}}&
\multicolumn{1}{c}{\makecell{$\text{NoC}_{10}$\\$@40$}}&
\multicolumn{1}{c}{\makecell{$\text{NoC}_{10}$\\$@50$}}&
\multicolumn{1}{c}{\makecell{$\text{FR}_{5}$\\$@6$}}&
\multicolumn{1}{c}{\makecell{$\text{NoC}_{5}$\\$@6$}}&
\multicolumn{1}{c}{\makecell{$\text{NoC}_{5}$\\$@8$}}&
\multicolumn{1}{c}{\makecell{$\text{NoC}_{5}$\\$@10$}}&
\multicolumn{1}{c}{\makecell{$\text{NoC}_{5}$\\$@12$}}\\
\midrule \midrule
 $1/8$  & max & sigmoid  & low $t_d, t_a$ 
&\blue{\textbf{3.91}}&\blue{1.94}&\blue{1.44}&\blue{\textbf{1.10}}&\blue{\textbf{0.89}} %225 combine3
&\blue{57.50}&\blue{\textbf{4.28}}&\blue{\textbf{3.20}}&\blue{\textbf{2.01}}&\blue{\textbf{1.31}}\\ %495
 $1/4$ & max & sigmoid  & low $t_d, t_a$ 
&\blue{7.03}&\blue{2.23}&\blue{1.54}&\blue{1.23}&\blue{1.04} %227 combine2
&\blue{56.25}& \blue{4.42}&\blue{3.31}&\blue{2.20}&\blue{1.38}\\ %?
 $1/2$ &  max & sigmoid & low $t_d, t_a$
&\blue{4.69}&\blue{\textbf{1.70}}&\blue{\textbf{1.30}}&\blue{1.12}&\blue{0.94} %224 combine1
&\blue{67.50}&\blue{4.44}&\blue{3.58}&\blue{2.24}&\blue{1.56}\\ %429
% \midrule
\cmidrule(l{1.5pt}r{1.5pt}){1-14}
 $1/8$ & average &sigmoid  & low $t_d, t_a$ 
&\blue{5.47}&\blue{2.30}&\blue{1.76}&\blue{1.39}&\blue{1.18} %432 average
&\blue{\textbf{55.00}}&\blue{4.41}&\blue{3.24}&\blue{2.25}&\blue{1.50} \\ %417
 $1/8$ &  max  &softmax & low $t_d, t_a$  
&\blue{8.59}&\blue{2.09}&\blue{1.70}&\blue{1.30}&\blue{1.12} %431 max
&\blue{70.00}&\blue{4.61}&\blue{3.56}&\blue{2.41}&\blue{1.65}\\ %420
\cmidrule(l{1.5pt}r{1.5pt}){1-14}
% \cmark & $1/8$ & \cmark & max &sigmoid  &\cmark &low $t_d, t_a$ & \\
 $1/8$ & max &sigmoid  &high $t_d, t_a$
&\blue{5.47}&\blue{2.03}&\blue{1.51}&\blue{1.21}&\blue{0.94} %446 high var
&\blue{77.50}&\blue{4.79}&\blue{3.85}&\blue{2.85}&\blue{1.98} \\ %424
 $1/8$ &  max &sigmoid  &adjacency 
&\blue{7.03}&\blue{2.34}&\blue{1.80}&\blue{1.40}&\blue{1.16} %454 adj
&\blue{75.00}&\blue{4.71}&\blue{3.73}&\blue{2.61}&\blue{1.73} \\ %426
% \cmidrule(l{1.5pt}r{1.5pt}){2-9}
% & max &sigmoid  &\cmark &low stdev. & - & \\
% \midrule
% \multirow{1}{*}{U-Net}
% & \cmark &  max & sigmoid  &\cmark&low variance & \\
\bottomrule
%%%%%%%%%%%%%%% Morph pairs
% & max &sigmoid& 5&5 &&\\
% & max &sigmoid&20&20&&\\
% \cmidrule(l{1.5pt}r{1.5pt}){2-8}
% & max &sigmoid&neighbor& 10&10 & &\\
%
\end{tabular}}
\end{center}
\label{table:sensitivitypublic}
\end{table*}

\begin{table}[t!]
\footnotesize
\caption{\textbf{Sensitivity analysis of the keypoint subset size for morphology-aware loss on the \spinewebdata dataset using \ourshort-v1.}
}
\vspace{-10pt}
\label{table:sensitivitymorph}
\begin{center}\resizebox{0.9\linewidth}{!}{%
\begin{tabular}{cc|ccccc}
\toprule
% \multicolumn{1}{c}{\multirow{2.5}{*}{Method}}&
\multicolumn{2}{c}{\multirow{1}{*}{\makecell{Subset size}}}&
% \multicolumn{1}{c}{\multirow{2.5}{*}{\makecell{Coordinate\\regression loss}}} &
\multicolumn{5}{c}{\multirow{1}{*}{Performance}} \\
\cmidrule(lr ){1-2}
\cmidrule(lr ){3-7}
% \multicolumn{1}{c}{} & 
\multicolumn{1}{c}{\makecell{$n(\mathcal{P}_d)$}} &
\multicolumn{1}{c}{\makecell{$n(\mathcal{P}_a)$}} &
% \multicolumn{1}{c}{} &
\multicolumn{1}{c}{\makecell{$\text{FR}_{10}$\\$@20$}}&
\multicolumn{1}{c}{\makecell{$\text{NoC}_{10}$\\$@20$}}&
\multicolumn{1}{c}{\makecell{$\text{NoC}_{10}$\\$@30$}}&
\multicolumn{1}{c}{\makecell{$\text{NoC}_{10}$\\$@40$}}&
\multicolumn{1}{c}{\makecell{$\text{NoC}_{10}$\\$@50$}}\\
\midrule \midrule
% {\multirow{9}{*}{\ourshort-v1}} & 
 40 & 40   & 9.38 &2.60 &2.02 & 1.56 &1.34 \\
 50 & 50   & 8.59 &2.18 &1.69 & 1.34 &1.15 \\
 60 & 60   & 7.81 &2.22 &\textbf{1.59} & \textbf{1.23} &\textbf{0.99} \\
 70 & 70   & \textbf{6.25} &2.46 &1.88 & 1.41 &1.19 \\
80 & 80  & 7.03 &2.88 &2.18 & 1.62 &1.21 \\
90 & 90  & 7.81 &2.20 &1.61 & 1.26 &1.01 \\
 100 & 100  &8.59 &2.35 &1.90 & 1.56&1.30 \\
 110 & 110 & \textbf{6.25} &\textbf{2.08} &\textbf{1.59} & 1.24 &1.06 \\
 120 & 120  & 8.59 &2.96 &2.27 & 1.63 &1.29 \\
 \midrule
 \xmark & \xmark &  12.50 & 3.68  &2.62 &1.94 &1.54 \\
\bottomrule
\end{tabular}}
\end{center}
\end{table}

\noindent
\textbf{Consistent superiority over baselines.}
\rbt{We present a detailed analysis of ARNet-v2's performance across varying target MRE thresholds and different user interaction rounds, as shown in Fig.~\ref{fig:noc}.
The results, as shown in Fig. 9a, demonstrate that ARNet-v2 consistently requires significantly fewer user clicks to reach target accuracy compared to all baselines, which require far more corrections to achieve the same performance. 
As illustrated in Fig. 9b, ARNet-v2 consistently achieves the lowest error over successive interaction rounds (Fig. 9b) compared to ARNet-v1 and the baseline models. This consistency highlights the robustness and reliability of our approach.
Furthermore, ARNet-v2 shows substantial error reductions in model revision (solid line) compared to manual revision (dashed line), demonstrating its effectiveness in leveraging user feedback.} Notably, models utilizing our interactive keypoint estimation framework, including ARNet-v2, ARNet-v1, and modified interactive segmentation models, display rapid error reduction (solid lines), in stark contrast to the much slower improvements observed with manual revisions (dashed lines). 
These results underscore the effectiveness of our proposed framework in enhancing annotation efficiency.
\rbt{In contrast, ClickPose exhibits negligible improvements over manual revisions, failing to refine additional keypoints after a user correction. Notably, as user hints increase, the performance of ClickPose falls below Dai et al., which does not even support user interaction. Additionally, \ourshort-v2 stands out for its lowest initial prediction accuracy, which is pivotal as it lays the groundwork for efficient subsequent refinements. }
These findings underscore the effectiveness of ARNet-v2 in minimizing user effort and achieving high accuracy by effectively propagating user interaction signals to all keypoints. 

\begin{figure*}[t!]
\begin{center}
\includegraphics[width=1.0\linewidth]{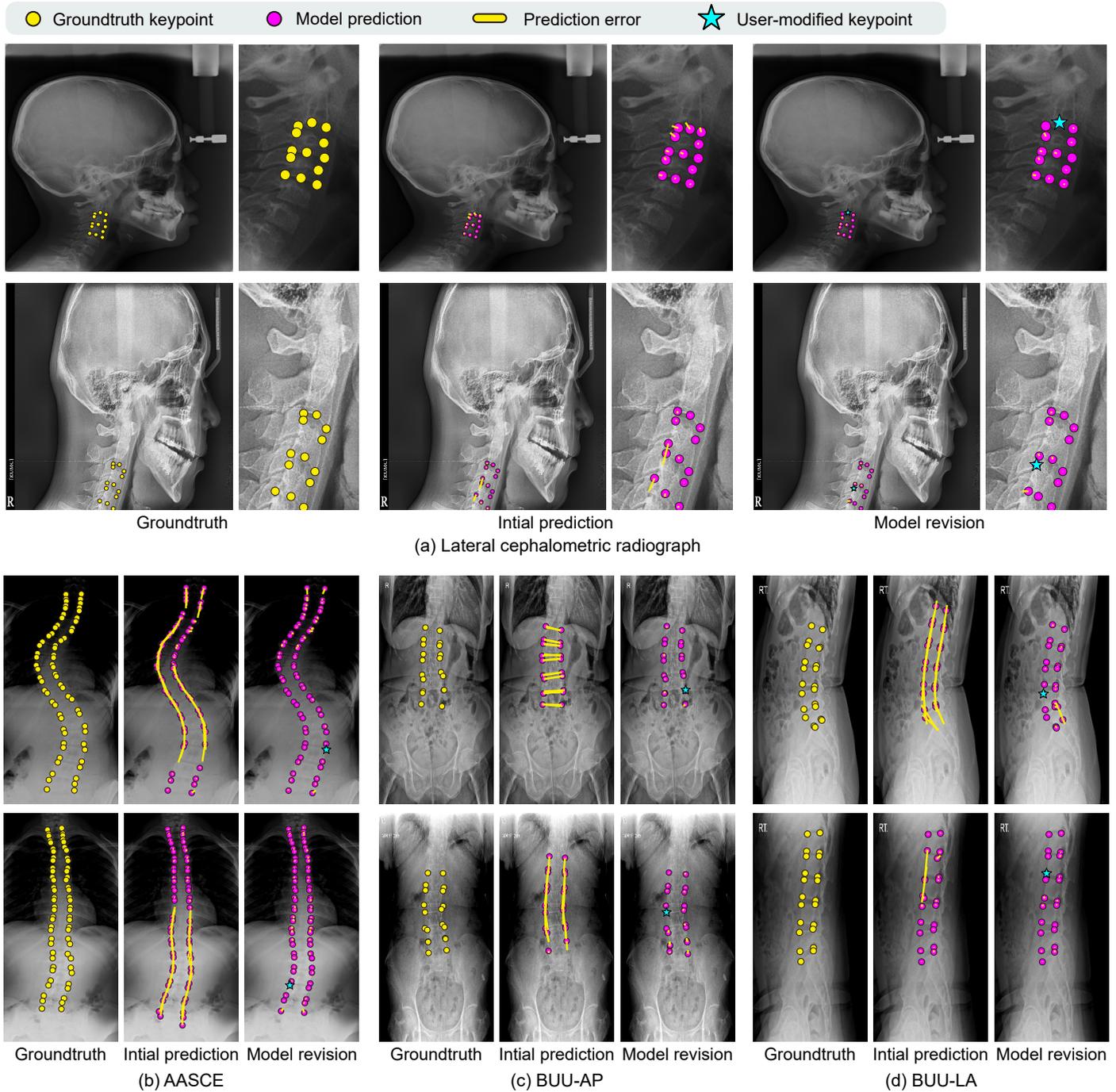}
\end{center}
\vspace{-10pt}
  \caption{\rbt{\textbf{Error refinement results of \ourshort-v2 after a single user correction.} 
Prediction errors are visualized by lines connecting each predicted keypoint to its corresponding groundtruth location. The length of these lines indicates the magnitude of the error, with shorter lines representing lower errors.}}
\label{fig:quality}
\end{figure*}

\subsubsection{Ablation study}
In our ablation study, we validate each component of our method, as shown in Table~\ref{table:ablation}. The results demonstrate that our complete model outperforms its ablated versions, highlighting the contribution of each individual component to the overall effectiveness of our approach.  

\rbt{We conduct additional analyses to quantitatively assess the effectiveness of the proposed morphology-aware loss in maintaining consistent inter-keypoint relationships. Specifically, we measure the error reduction in angles and distances between consistent keypoint pairs on the BUU-AP dataset when the morphology-aware loss is used in ARNet-v2. Following the metrics used in the loss itself, we calculate the L1 distance for inter-keypoint distances and cosine distance for inter-keypoint angles.}
\rbt{Our results show that the morphology-aware loss reduces errors in inter-keypoint distances by 17.1\% and errors in inter-keypoint angles by 47.4\% compared to the model without this loss. These findings quantitatively confirm that the morphology-aware loss is highly effective in preserving consistent inter-keypoint relationships, ensuring higher anatomical consistency.}

\rbt{This improvement directly enhances keypoint prediction performance. As demonstrated in Table~\ref{table:ablation}, incorporating the morphology-aware loss reduces the failure rate of ARNet-v2 on the AASCE dataset from 7.81\% to 3.91\%. Similarly, on the BUU-AP dataset, the failure rate decreases from 60\% to 57.5\%. Moreover, the number of required user clicks (NoC) consistently decreases when the morphology-aware loss is applied, highlighting its effectiveness in error correction by preserving inter-keypoint relationships.}
\rbt{The effectiveness of the morphology-aware loss is also evident in ARNet-v1. As shown in Table~\ref{table:sensitivitymorph}, incorporating this loss reduces the failure rate from 12.5\% to 6.25\%, while also consistently lowering NoC. These improvements demonstrate how the morphology-aware loss enhances keypoint prediction accuracy by preserving inter-keypoint anatomical consistency.}

\subsubsection{Sensitivity analysis}
We conduct a thorough sensitivity analysis of \ourshort to see how various configurations impact model performance, as detailed in Table~\ref{table:sensitivitypublic}. 
First, we investigate the impact of varying the downsampling scales of the feature map $F_u$ within the cross-attention mechanism of \attendOurs (Fig.~\ref{fig:arnet}c). Experimenting with downsampling ratios of $\frac{1}{8}$, $\frac{1}{4}$, and $\frac{1}{2}$, the result underlines the robustness of our model across a range of resolutions.
Second, in the gating operation of \attendOurs (Fig.~\ref{fig:arnet}b), we explore different pooling methods (average vs. max pooling) and activation functions (sigmoid vs. softmax). This analysis assesses how feature aggregation and activation choices influence keypoint refinement performance. We find that global max pooling combined with a sigmoid activation function yields better results.

Lastly, we explore the impact of altering the distance and angle set criteria within our morphology-aware loss. This investigation highlights the importance of these criteria for effectively capturing inter-keypoint relationships. Our findings indicate that consistent relationships notably improve model performance, surpassing outcomes from either highly variable ones or adjacent relationships. 
Additionally, we showcase results obtained using different threshold values in the morphology-aware loss, as shown in Table~\ref{table:sensitivitymorph}.
These findings demonstrate the necessity of careful threshold selection to balance the flexibility and stability of keypoint relationships.

\subsubsection{Qualitative results on interactive keypoint estimation}
We present qualitative results of model predictions across all four datasets, as illustrated in Fig.~\ref{fig:quality}. 
This comparison includes both initial predictions and the outcomes following one round of user revision. 
A standout observation from this analysis is the remarkable efficiency of \ourshort-v2 in accurately positioning multiple keypoints with just one user modification.
This highlights the model's capability to effectively propagate user feedback, even to keypoints that are far from the area of direct user interaction.
The significance of these findings lies in their practical implications: \ourshort-v2 substantially reduces the manual effort traditionally required in the keypoint annotation process. By eliminating the need for users to individually adjust each keypoint, our method streamlines the annotation process, making it more efficient and less time-consuming. This efficiency not only enhances the user experience but also ensures greater accuracy and consistency in the results, reinforcing the practical utility of \ourshort-v2 in diverse application scenarios.

\begin{figure*}[t!]
\begin{center}  \includegraphics[width=0.9\linewidth]{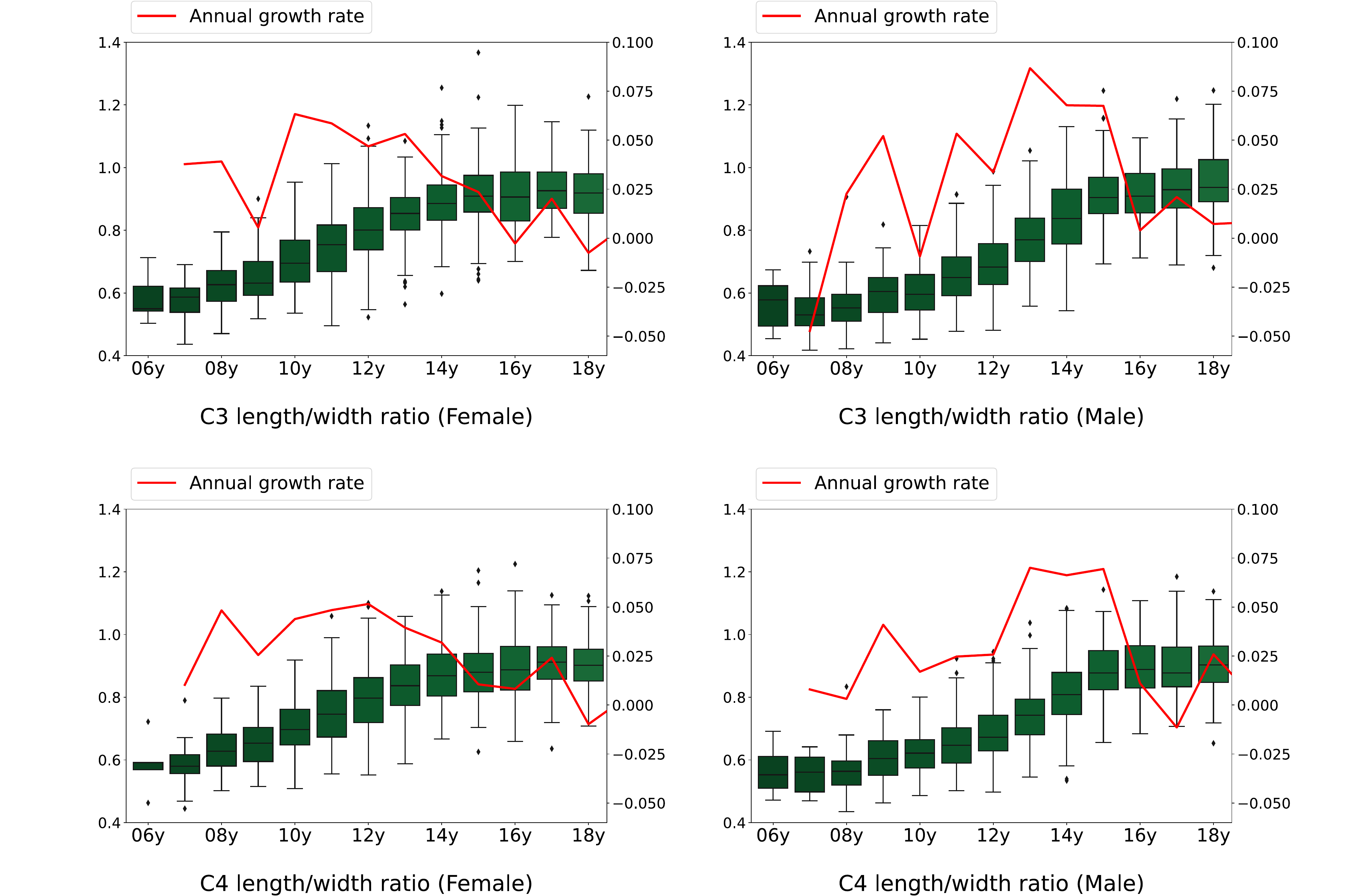}
\end{center}
\vspace{-10pt}
  \caption{\textbf{Annual growth rates based on standard growth curves.} The results highlight the growth peak for each cervical vertebra. The complete results are available in Figs.~\ref{suppe_fig:diff1} and \ref{suppe_fig:diff2} in the Appendix~\ref{appen_sec:diff}. \rbtthree{The left y-axis indicates the CVM feature value, while the right y-axis represents the annual growth rate.}}
\label{fig:diff}
\end{figure*}

\begin{table*}[t!]
    \caption{\textbf{Cervical vertebrae morphology at growth peak.} It outlines detailed ratios indicating peak growth stages.}\label{tab:difference}
    \begin{center}
    \vspace{-10pt}
    \resizebox{0.7\linewidth}{!}{%
                \begin{tabular}{c|cc|cc|cc|cc|cc|cc}
    \toprule
          \multirow{2.5}{*}{Sex}
         & \multicolumn{6}{c}{C3 length/width ratio} 
         & \multicolumn{6}{c}{C4 length/width ratio}\\
         \cmidrule(lr ){2-7} \cmidrule(lr ){8-13}
         & Age & Value & Age  & Value & Age  & Value 
         & Age & Value & Age  & Value & Age  & Value \\
         \midrule
         \midrule
         Female
         & 9y & 0.63 & \textbf{10y} & \textbf{0.70} & 11y & 0.75 
         & 9y & 0.65 & \textbf{10y} & \textbf{0.70} & 11y & 0.75\\
         Male
         & 12y & 0.68 & \textbf{13y} & \textbf{0.77} & 14y & 0.84 
         & 12y & 0.67 & \textbf{13y} & \textbf{0.74} & 14y & 0.81 \\
     \bottomrule
    \end{tabular}}
\end{center}
\end{table*}

\subsubsection{Growth peak in cervical vertebrae}\label{sec:results_growth}
Our study delves into the growth rates of the third and fourth cervical vertebrae (C3 and C4), specifically examining changes in their length/width ratio.
As illustrated in Fig.~\ref{fig:diff}, we observe significant growth peaks in these ratios at age 10 for females and 13 for males, indicating sex-specific growth patterns in the cervical vertebrae.
The periods one year before and after these ages (9 to 11 for females and 12 to 14 for males) are critical for orthodontic interventions, as a patient's remaining growth potential can be fully utilized to maximize the effects of dentofacial orthopedic treatment.

We find that at these growth peaks, the C3 length/width ratio reaches about 0.70 for females and 0.77 for males, as detailed in Table~\ref{tab:difference}. 
These ratios are indicative of cervical vertebrae morphology at the growth peak.
This is also applicable to C4; when the length of C4 reaches about 70\% of the width for females and 74\% for males, it suggests that the growth rate reaches the highest phases.
That is, patients exhibiting these specific ratios are likely in their highest growth rate phases.

\begin{figure}[t!]
\begin{center}  \includegraphics[width=0.77\linewidth]{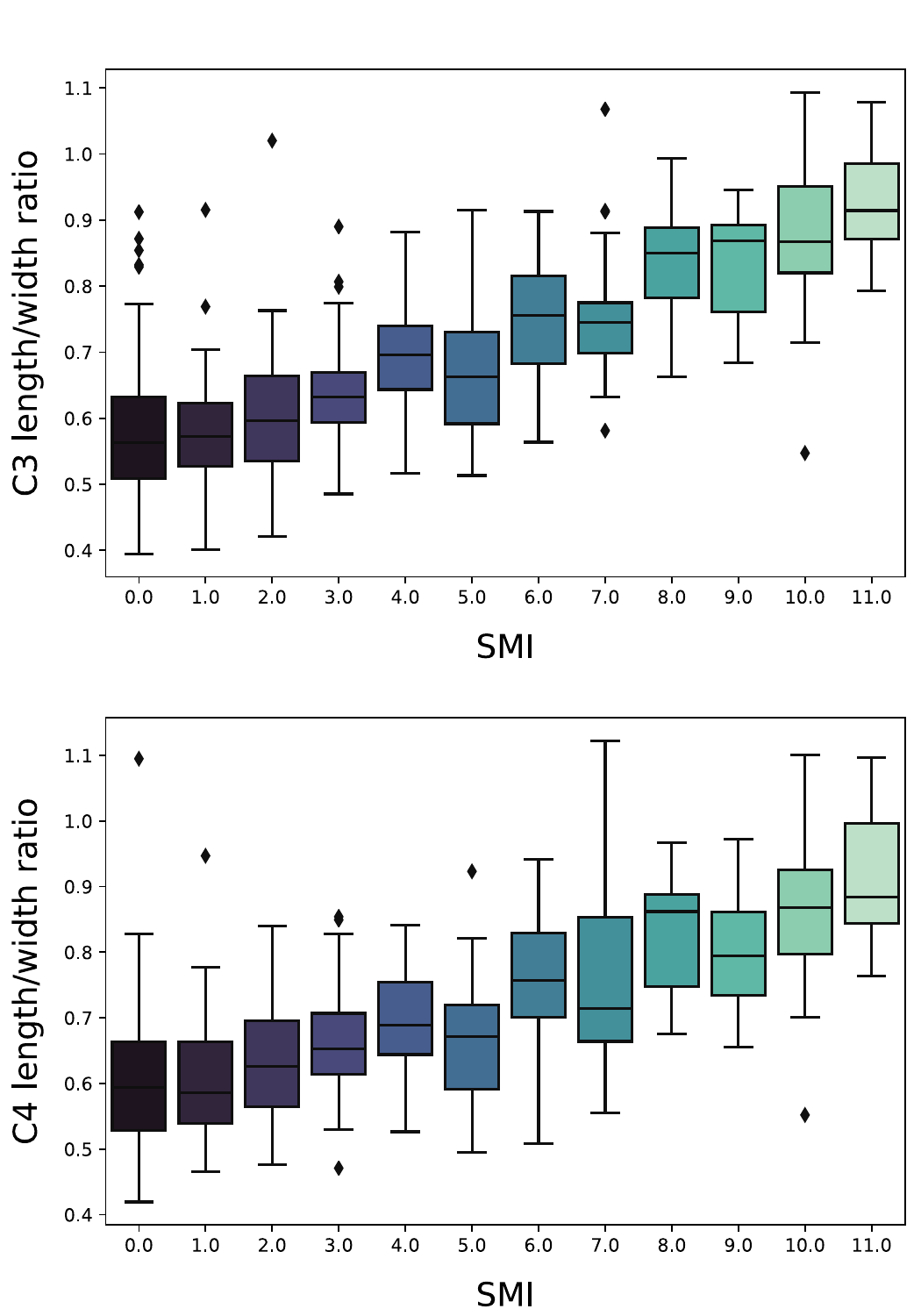}
\end{center}
\vspace{-10pt}
  \caption{\textbf{Correlation between SMI and the C3 and C4 length/width ratios.}}
\label{fig:smi_cvm}
\end{figure}

Our analysis further investigates the relationship between CVM features and SMI from hand-wrist radiographs. The Pearson correlation coefficients between length/width ratio and SMI are 0.77 for C3 and 0.72 for C4. The peak growth velocity for both C3 and C4 correspond with the SMI 4 stage, indicating length/width ratios of these vertebrae as reliable diagnostic references, as illustrated in Fig~\ref{fig:smi_cvm}.

These findings hold significant clinical implications for orthodontic practice. They provide a detailed reference for assessing a patient's remaining growth potential, which is essential for planning effective treatments. 
Aligning treatment with these cervical vertebrae morphology at growth peak allows clinicians to fully leverage the patient's growth potential during treatment, thereby maximizing treatment efficacy.

\subsubsection{{Impact of keypoint errors on growth peak analysis}}\label{exp:errorpropa}

\rbt{We analyze the impact of keypoint location errors on CVM feature calculations and their potential influence on growth peak assessment. Specifically, we examine how keypoint errors affect the C3 length/width ratio and, consequently, the timing of the growth peak. Furthermore, we introduce an evaluation method to assess the performance of keypoint estimation models based on their impact on clinical analysis.}

\rbt{Using the lateral cephalometric radiograph dataset, we measure the average changes in CVM features when a single keypoint among the five points defining C3 is shifted by 1 pixel in any direction (up, down, left, right) within a 512×512 image. Our analysis reveals that a 1-pixel keypoint error causes an average absolute error of 0.0127 in the C3 length/width ratio.}

\rbt{To quantify how variations in the C3 length/width ratio influence growth peak analysis, we examine the CVM value boundaries at the growth peak. As shown in Table~\ref{tab:difference}, for females, the boundaries for the C3 length/width ratio at the growth peak are 0.63–0.70 and 0.70–0.75, with central values at 0.665 and 0.725, respectively. A deviation from the growth peak ratio of 0.70 by -0.035 or +0.025 would lead to an incorrect growth peak estimate. This corresponds to a 3.57\% error in the C3 length/width ratio for females. Similarly, for males, the growth peak ratio is 0.77, with a tolerance of -0.045 or +0.035, corresponding to a 4.55\% error. 
Thus, a deviation of approximately 0.025 in the ratio can alter growth peak timing predictions.} 

\rbt{Combining these findings, we estimate that a 1.97$\approx$2-pixel keypoint error would result in a 0.025 error in the C3 length/width ratio, sufficient to affect growth peak analysis.
In contrast, the MRE of keypoint predictions of ARNet-v2 is 1.163 pixels, resulting in an average C3 length/width ratio error of 0.0148. This error is below the threshold required to affect growth peak timing predictions, indicating that ARNet-v2 maintains sufficient accuracy for this application.}

\rbt{To evaluate the clinically relevant errors of keypoint estimation models, we introduce a failure rate metric based on the threshold at which keypoint errors influence growth peak predictions. Specifically, we define failure as cases where the keypoint error exceeds 2 pixels, sufficient to alter growth peak timing as described above. Using this threshold, we evaluate the percentage of images in which keypoint prediction exceeds the 2-pixel error threshold.}
\rbt{The results, as shown in Table~\ref{table:initial_failure_rate_FR}, demonstrate that ARNet-v2 achieves the lowest failure rate at 1.88\%, followed closely by ARNet-v1 at 1.97\%. In comparison, ClickPose and the model proposed by Dai et al. exhibit significantly higher failure rates of 99.9\% and 30.23\%, respectively. These findings indicate that competing keypoint estimation models frequently produce keypoint errors large enough to impact growth peak timing predictions, while ARNet-v2 maintains superior accuracy in this clinically relevant context. These analyses demonstrate that ARNet-v2 maintains low clinically relevant errors and outperforms baseline models by a substantial margin.
}

% \begin{table}[t!]
% \caption{\rbt{
%  Comparison of failure rates in achieving the target MRE of 2 pixels with initial keypoint estimation on the Lateral cephalometric radiograph dataset.
% }}
% \label{table:initial_failure_rate_FR}
% \vspace{-10pt}
% \begin{center}
% \resizebox{1.0\linewidth}{!}{%
% \begin{tabular}{c|ccccccc}
% \toprule
% % ===== 맨위 ===========
% \rbt{Method} & \rbt{BRS} & \rbt{f-BRS} & \rbt{RITM} & \rbt{ClickPose} & \rbt{Dai et al.} & \rbt{ARNet-v1} & \rbt{ARNet-v2}  \\ 
% % ================== 이제 성능 기록 ========
% \midrule\midrule
% \rbt{$\text{FR}_0$@2 (pixel)} & \rbt{5.00} & \rbt{1.90} & \rbt{3.64} & \rbt{99.91} & \rbt{30.23} & \rbt{1.97} & \rbt{\textbf{1.88}}\\
% \bottomrule
% \end{tabular}}
% \end{center}
% \end{table}

% \begin{table}[t!]
% \caption{\rbt{
%  Comparison of failure rates in achieving the target MRE of 2 pixels with initial keypoint estimation on the Lateral cephalometric radiograph dataset.
% }}
% \label{table:initial_failure_rate_FR}
% \small
% \vspace{-10pt}
% \begin{center}
% % \resizebox{1.0\linewidth}{!}{%
% \begin{tabular}{l|c}
% \toprule
% % ===== 맨위 ===========
% \multicolumn{1}{c}{\rbt{Method}} & \rbt{$\text{FR}_0$@2 (pixel)}  \\ 
% \midrule\midrule
% % ================== 이제 성능 기록 ========
% \rbt{BRS} & \rbt{5.00} \\
% \rbt{f-BRS} & \rbt{1.90} \\
% \rbt{RITM} & \rbt{3.64} \\
% \rbt{ClickPose} & \rbt{99.91} \\
% \rbt{Dai et al.} & \rbt{30.23} \\
% \rbt{ARNet-v1} & \rbt{1.97} \\
% \rbt{ARNet-v2} & \rbt{1.88} \\
% \bottomrule
% \end{tabular}
% \end{center}
% \end{table}

\begin{table}[t!]
\caption{\rbt{
 \textbf{Comparison of failure rates of keypoint estimation models in impacting the growth peak analysis results.} Failure rates are measured for the target MRE thresholds ranging from 1 pixel to 2 pixels, based on initial keypoint estimation on the Lateral cephalometric radiograph dataset.
}}
\label{table:initial_failure_rate_FR}
\small
\vspace{-10pt}
\begin{center}
\resizebox{1.0\linewidth}{!}{%
\begin{tabular}{l|cccccc}
\toprule
% ===== 맨위 ===========
\multicolumn{1}{c}{\rbt{Method}} & 
    \multicolumn{1}{c}{\makecell{\rbt{$\text{FR}_0$}\\\rbt{$@1.0$}}}  &  
    \multicolumn{1}{c}{\makecell{\rbt{$\text{FR}_0$}\\\rbt{$@1.2$}}}  & 
    \multicolumn{1}{c}{\makecell{\rbt{$\text{FR}_0$}\\\rbt{$@1.4$}}}  & 
    \multicolumn{1}{c}{\makecell{\rbt{$\text{FR}_0$}\\\rbt{$@1.6$}}}  & 
    \multicolumn{1}{c}{\makecell{\rbt{$\text{FR}_0$}\\\rbt{$@1.8$}}}  &
    \multicolumn{1}{c}{\makecell{\rbt{$\text{FR}_0$}\\\rbt{$@2.0$}}}  
    \\ 
\midrule\midrule
% ================== 이제 성능 기록 ========
\rbt{BRS} & \rbt{83.62} & \rbt{57.00} & \rbt{31.70} & \rbt{16.49} & \rbt{8.48} & \rbt{5.00} \\
\rbt{f-BRS} & \rbt{65.14} & \rbt{34.45} & \rbt{16.04} & \rbt{6.95} & \rbt{3.72} & \rbt{1.90} \\
\rbt{RITM} & \rbt{80.59} & \rbt{50.85} & \rbt{27.09} & \rbt{13.33} & \rbt{6.84} & \rbt{3.64} \\
\rbt{ClickPose} & \rbt{100.0} & \rbt{100.0} & \rbt{100.0} & \rbt{100.0} & \rbt{99.98} & \rbt{99.91} \\
\rbt{Dai et al.} & \rbt{99.63} & \rbt{96.34} & \rbt{86.41} & \rbt{67.13} & \rbt{46.01} & \rbt{30.23} \\
\midrule
\rbt{ARNet-v1} & \rbt{65.57} & \rbt{34.19} & \rbt{15.36} & \rbt{6.82} & \rbt{3.46} & \rbt{1.97} \\
\rbt{ARNet-v2} & \rbt{\textbf{61.13}} & \rbt{\textbf{30.86}} & \rbt{\textbf{13.48}} & \rbt{\textbf{6.21}} & \rbt{\textbf{3.35}} & \rbt{\textbf{1.88}} \\
\bottomrule
\end{tabular}}
\end{center}
\end{table}

\subsubsection{Individual variations in cervical vertebrae growth rate}
We have shown the median length/width ratio curves of C3 and C4 with the annual growth rate in our sample. However, it should be noted that there is an individual variability in the growth rate and the timing of the pubertal peak. Therefore, we randomly selected patients who have more than three serial lateral cephalometric radiographs and analyzed individual growth rates using the length/width ratio values of C3 and C4, and compared them with the median growth rate curve.
By examining the growth rates alongside the median growth rates, as depicted in Fig.~\ref{fig:cvm_for_each_patient}, we provide a clear visual representation of the variability in CVM features among individuals. 
 This visualization distinctly illustrates the different timings in pubertal growth peak in the CVM features, which are closely linked to the remaining growth potential for each individual.

\section{Limitation and future work}
One limitation of this study is that errors in keypoint estimation can affect the accuracy of growth potential estimations, \rbt{as discussed in Section~\ref{exp:errorpropa}}. This highlights the need for further refinement of deep learning models to reduce these errors and enhance the overall precision of growth potential estimation.

Additionally, similar to the challenges faced in previous work~\citep{baccetti2002improved,g9}, our research is limited by the lack of substantial longitudinal data for individual patients. Only 180 patients in our datasets have longitudinal data, with serial radiographs collected over periods ranging from 2 to 10 years. This limited sample size, combined with large heterogeneity within each age group, results in a wide range of CVM feature values and growth curves with restricted generalizability. For example, while skeletal growth is generally understood to exhibit positive growth rates, indicating an increase in CVM features with age until 18, our analysis occasionally reveals negative growth rates due to the cross-sectional analysis of the age groups.
Nonetheless, the overall pattern of acceleration in growth velocity
until the pubertal peak, followed by a subsequent deceleration, is clearly observed in our growth curves.

\begin{figure}[t!]
\begin{center}  \includegraphics[width=1.0\linewidth]{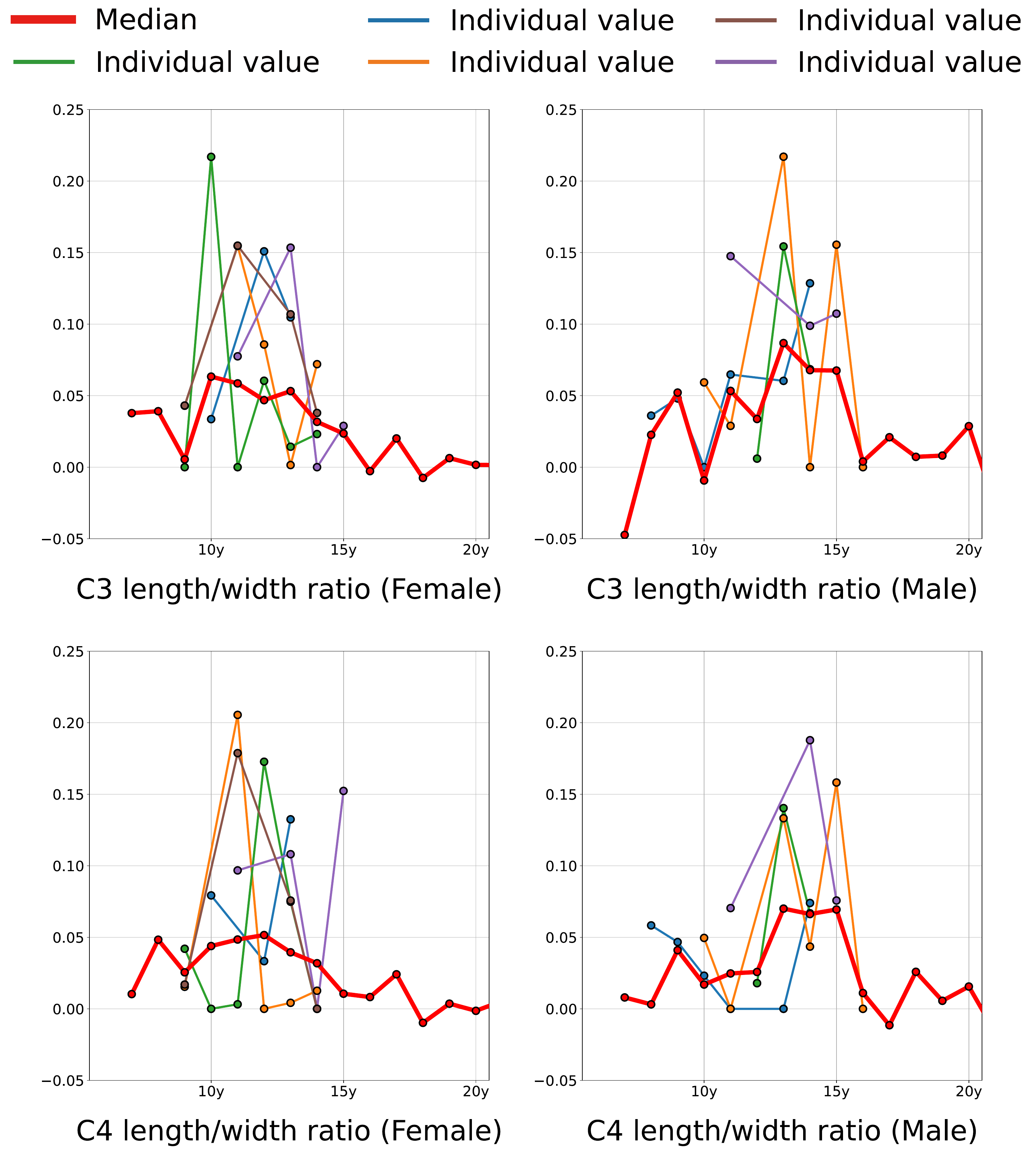}
\end{center}
\vspace{-10pt}
  \caption{\textbf{Individual growth rates alongside median growth rate.} For each sex, the growth rate of each patient is depicted in the same color for C3 and C4.}
\label{fig:cvm_for_each_patient}
\end{figure}

\rbt{To capture cervical vertebrae growth trends, we analyze growth peak timing using median values for each age group, capturing cervical vertebrae growth patterns at the population level. Additionally, we examine individual growth patterns, offering insights into specific growth trajectories, as illustrated in Fig.~\ref{fig:cvm_for_each_patient}. While the use of the median offers advantages at the population level, we acknowledge that growth patterns vary among individuals. The lack of substantial longitudinal data for individual patients limits our ability to fully capture this variability, particularly for atypical growth patterns. Additionally, as a limitation of a retrospective study, our dataset lacks annual records for patients with longitudinal data, restricting our ability to precisely determine the timing of peak growth velocity. These limitations may introduce potential errors in interpreting individual variability in annual growth patterns.}
\rbt{Future prospective studies incorporating substantial longitudinal data from individual patients would enhance the robustness of these findings. Such studies could validate the accuracy of growth potential predictions and provide a more comprehensive understanding of individual growth trajectories. While beyond the scope of this work, this represents a critical direction for future research. }

Lastly, while it is widely accepted that the skeletal maturity of the cervical vertebrae is closely associated with the growth of the maxilla and the mandible, this study focuses solely on the growth of the cervical vertebrae without directly comparing it to jaw growth. 
Further research should explore the correlation between the growth rates of the cervical vertebrae and the jaws, as understanding this relationship would provide more reliable guidelines for comprehensive orthodontic diagnosis and treatment planning.
Furthermore, expanding the sample size to include a more diverse range of patients across various ages and ethnicities will increase the generalizability of the results.

Despite these limitations, \rbt{our study provides valuable insights into general quantitative changes in cervical morphological characteristics during the growth phase, which has been underexplored in prior research. These findings offer clinically valuable guidance for determining the optimal timing for orthodontic treatments,} representing a significant contribution to the field of clinical orthodontics. It underscores the importance of accurate growth potential estimation in orthodontic treatment planning and sets the stage for further advancements in this area. By addressing these limitations in future research, we can continue to improve the effectiveness and efficiency of orthodontic treatments, ultimately benefiting clinical practice.

\section{\rbt{Broader applicability}}
\rbt{ARNet-v2 offers several advantages that make it highly adaptable and effective for broader medical image annotation tasks.}

\rbt{The cross-attention mechanism in ARNet-v2 is particularly effective for annotating spatially distant keypoints, offering significant benefits for high-resolution medical images such as CT or MRI scans, where efficient identification and utilization of critical signals are essential. By selectively retrieving user interaction signals at the pixel level, ARNet-v2 ensures accurate estimation of keypoints even across wide spatial areas, as shown in Fig.~\ref{fig:Arnet2_intro}.}
\rbt{Additionally, the ability of ARNet-v2 to handle a wide range of keypoints, from 13 to 68 keypoints in our study, establishes it as a versatile tool for tasks requiring extensive keypoint annotations. Additionally, our results demonstrate that ARNet-v2 performs robustly on images from both lateral and anterior-posterior views, underscoring its capability to function effectively across various imaging perspectives and making it suitable for diverse medical annotation tasks.}

\rbt{ARNet-v2 also significantly enhances annotation efficiency by improving initial prediction accuracy and enabling precise error correction through the effective propagation of user interaction signals. This results in high precision with minimal user clicks, leading to faster and more efficient annotations. By reducing the time and effort required for manual annotation, ARNet-v2 allows clinicians to focus more on interpreting results and improving patient care outcomes. }
\rbt{Furthermore, the increased efficiency supports scalability in dataset collection, facilitating the creation of larger, high-quality datasets for future studies. This scalability enables broader exploration and contributions across various research domains, further enhancing the impact of ARNet-v2.  }

\rbt{These strengths in accuracy, efficiency, scalability, and adaptability demonstrate the potential of ARNet-v2 to support a wide range of medical image annotation tasks. Extending ARNet-v2 to additional annotation tasks would further advance clinical research and enhance patient care, highlighting its versatility and broad impact in addressing challenges in the field.
}

\section{Conclusions}
The ability to accurately estimate bone age and predict the remaining growth potential during the crucial developmental stages of childhood and adolescence is of immense importance in the field of orthodontics, particularly for diagnosis and treatment planning. Our study contributes significantly to this area by enabling precise predictions of a patient's pubertal growth peak based on their current cervical vertebra morphology. By analyzing cervical vertebral maturation (CVM) feature values, we can determine the morphological characteristics at the growth peak, which are key predictors of growth potential. This approach provides invaluable diagnostic guidance and assists clinicians in identifying the optimal timing for initiating dentofacial orthopedic treatments.
Furthermore, we propose a novel interactive keypoint estimation network, referred to as \ourshort. This network integrates advanced components, such as the \attendOurs mechanism and morphology-aware loss function, to facilitate accurate keypoint revision with minimal manual intervention. Our comprehensive experiments and analyses, conducted across four medical datasets, have demonstrated the efficacy of this approach. 
The ability to simultaneously correct numerous keypoints accurately with minimal user input underscores the practical utility and advanced capabilities of our model in streamlining the keypoint annotation process.

% \section*{Acknowledgments}
% Acknowledgments should be inserted at the end of the paper, before the
% references, not as a footnote to the title. Use the unnumbered
% Acknowledgements Head style for the Acknowledgments heading.

% \section*{References}
%%Harvard
\bibliographystyle{model2-names.bst}\biboptions{authoryear}
\bibliography{refs}

\newpage
\appendix
\renewcommand\thesection{\Alph{section}}
\setcounter{figure}{0}
\setcounter{table}{0}

\section*{Supplementary Material}
This supplementary section provides additional materials to complement the main manuscript. These include detailed visualizations and extended experimental results, which enrich the understanding of our study.
\setlist{itemsep=0.5pt,topsep=3pt}
\begin{itemize}
\item 
    \textbf{Appendix~\ref{appen_sec:notation}} describes a comprehensive description of all notations used in the main manuscript.
\item 
    \rbt{\textbf{Appendix~\ref{appen_sec:baseline_reproducibility}} provides implementation details of the baseline models.}
\item 
\rbttwo{
\textbf{Appendix~\ref{appen_sec:labelefficient}} details similarities and differences of our method with label-efficient methods.
}

\item 
    \textbf{Appendix~\ref{appen_sec:demo}} shows an example of using our web-based demo tool to highlight its practical application in orthodontic and orthopedic planning. 

\item 
    \textbf{Appendix~\ref{appen_sec:qual}} illustrates a qualitative comparison of \ourshort-v1 and the baseline models.
\item 
    \textbf{Appendix~\ref{appen_sec:diff}} provides complete standard growth curves and growth rates, including results for all seven cervical vertebral maturation features.

\end{itemize}

\section{Notation summary}\label{appen_sec:notation}
We provide a detailed description of all the notations used throughout our paper in Table~\ref{table:notation}. Each notation is defined with its corresponding dimension and a clear description, facilitating easy reference and a better understanding of our work.

\begin{table*}[!ht]
\caption{\textbf{Summary of notations.}}
\label{table:notation}
\vspace{-10pt}
\begin{center}
\resizebox{0.85\textwidth}{!}{%
\begin{tabular}{c|c|l}
\toprule
% {Notation}
Notation&
{Dimension}&
{Description}\\
\midrule
\midrule
\multicolumn{3}{c}{\ourmethod}\\
\midrule
$C$, $W$, $H$ & $\mathds{R}^{1}$&  
channel dimension, width, height of an input image \\
 $K$ & $\mathds{R}^1$&  
 total number of keypoints \\
 $\mathcal{I} $ & $\mathds{R}^{C\times W \times H}$ &
 input image\\
  $\mathcal{U}$ & $\mathds{R}^{K\times W \times H}$ &
 user-interaction heatmap\\
$\mathcal{H}$& $\mathds{R}^{K\times W \times H}$& 
groundtruth keypoint heatmaps  \\
$\hat{\mathcal{H}} $& $\mathds{R}^{K\times W \times H}$& 
 predicted keypoint heatmaps  \\
  $\sigma_u^2$ &  $\mathds{R}^{1}$&
 covariance of the Gaussian-smoothed user-interaction heatmap $\mathcal{U}$ \\
  $\sigma_h^2$ &  $\mathds{R}^{1}$&
 covariance of the Gaussian-smoothed groundtruth keypoint heatmap $\mathcal{H}$  \\
 $p_n$ &  $\mathds{R}^{2}$ &
groundtruth coordinates of the $n$-th keypoint; $p_n=(x_n,y_n)$ \\
  $\hat{p}_n$&  $\mathds{R}^{2}$ &
 predicted coordinates of the $n$-th keypoint \\
 $\hat{q}_n$&  $\mathds{R}^{2}$ & coordinate with the highest probability value in $\hat{\mathcal{H}}$\\
\midrule
\multicolumn{3}{c}{\attendOurs}\\
\midrule
 $k_{u}$, $k_c$, $k_s$ & $\mathds{R}^{1}$ & channel dimensions of intermediate feature maps\\
 $\textbf{F}_i$& $\mathds{R}^{k_{{u}} \times \frac{W}{4} \times \frac{H}{4} } $&
 intermediate feature map of the main network \\
  $\textbf{F}_u$& $\mathds{R}^{k_{{u}} \times \frac{W}{8} \times \frac{H}{8} } $& user-interaction-enriched feature map \\
 $\textbf{F}_c$& $\mathds{R}^{k_{c} \times \frac{W}{4} \times \frac{H}{4} }$ & original image feature map to recalibrate\\
 $\textbf{F}_a$ &$\mathds{R}^{k_{c} \times \frac{W}{4} \times \frac{H}{4} }$ & recalibrated image feature map  \\
   $\textbf{A}$ & $\mathds{R}^{k_c}$&gating weight  \\
   $w$, $h$ & $\mathds{R}^{1}$&  
width, height of a down sampled feature map for cross-attention \\
\midrule
\multicolumn{3}{c}{Morphology-aware loss} \\
\midrule
$ d_{m,n} $&$\mathds{R}^{1}$ &groundtruth distances between two keypoints, $p_m$ and $p_n$\\ 
$ \hat{d}_{m,n} $&$ \mathds{R}^{1}$ &predicted distances between two keypoints, $\hat{p}_m$ and $\hat{p}_n$\\ 
$ u_{m,n,l}$&$ \mathds{R}^{2}$& groundtruth angle vectors between three keypoints, $p_m$ and $p_n$, and $p_l$; $u=[u_x, u_y]$\\
$ \hat{u}_{m,n,l} $&$ \mathds{R}^{2}$& predicted angle vectors between three keypoints\\ 
$S_d, S_a$&$\mathds{R}^{1}$&standard deviation values of distance $d$, angle vector $u$\\
${t}_d, {t}_a$& $\mathds{R}^{1}$ & threshold values for $S_d, S_a$ to determine $\mathcal{P}_d, \mathcal{P}_a$\\
$\mathcal{P}_d, \mathcal{P}_a$&-& distance, angle sets to apply the proposed loss\\
$\lambda_a$, $\lambda_m$ & $\mathds{R}^{1}$& loss coefficients that range from zero to one. \\
$L_g$ & $\mathds{R}^{1}$& binary cross-entropy loss for keypoint heatmap\\
$L_d$ & $\mathds{R}^{1}$&  L1 loss for distance\\
$L_a$ &$\mathds{R}^{1}$ &  cosine similarity loss for angle vector \\
$L_m=L_d+\lambda_a L_a$ & $\mathds{R}^{1}$ & morphology-aware loss \\
$L=L_g+\lambda_m L_m$ &$\mathds{R}^{1}$ &  total loss\\
\bottomrule
\end{tabular}}
\end{center}
\end{table*}

\section{\rbt{Baseline reproducibility}}\label{appen_sec:baseline_reproducibility}
\noindent
\textbf{Keypoint estimation models}
\rbt{ClickPose, originally proposed for the multi-person keypoint detection task, assumes the presence of multiple instances in its predicted results. However, as this study focuses on single X-ray images, we modify ClickPose to predict only a single object per image. Additionally, the object keypoint similarity loss, specifically designed for multi-person keypoint detection, is excluded from this study.
The model proposed by Dai et al. is reimplemented using their official code. Since this model does not include an interactive module, we evaluate its performance by measuring the prediction accuracy after manually revising its initial predictions.}

\noindent
\textbf{Interactive segmentation models} \rbt{In this study, we implement interactive segmentation models, including BRS, f-BRS, and RITM, using their official source codes. The hyperparameters for these models are derived from the values used in our previous work.
The interactive segmentation models used as baselines share similar concepts with our proposed interactive keypoint estimation framework in the context of interactive systems. Both approaches leverage user interaction information to enhance model accuracy in predicting targets within an image. Consequently, we adopt shared metrics, such as the number of user clicks (NoC) and failure rates (FR). Moreover, both frameworks predict target locations at the pixel level using a probability heatmap. These make interactive segmentation models a strong baseline for interactive keypoint estimation, particularly in the absence of existing interactive keypoint models.}

\rbt{However, key differences between the two approaches necessitate adaptations when applying segmentation models to keypoint estimation. These modifications include adjustments to the user interaction representation, output layers, and loss function. The goal is to transform a model designed for image segmentation into one capable of accurately predicting discrete keypoints. }

\begin{itemize}
\item \rbt{\textbf{User interaction}: In segmentation-based baseline models, user interactions are encoded as hint heatmaps with two channels: one for positive clicks and another for negative clicks. For interactive keypoint estimation, however, there are no negative clicks, and each user input directly corresponds to the correct keypoint. To accommodate this, we adjust the hint heatmap representation to have the number of channels equal the number of keypoints, following our approach.}

\item \rbt{\textbf{Output layer}: For keypoint estimation, the model outputs a set of heatmaps, each corresponding to a keypoint, where the value of each pixel represents the likelihood of being the location of the target keypoint. We adapt the output layers of segmentation models to generate the required number of heatmaps and apply a sigmoid activation function.}

\item \rbt{\textbf{Loss function}: Keypoint estimation requires pixel-wise probability maps indicating the likelihood of keypoint locations. We use binary cross-entropy loss for each pixel, comparing predicted heatmaps to ground truth heatmaps.} 

\rbt{Additionally, for BRS and f-BRS, we observe that their backpropagation refinement schemes, though effective for segmentation tasks, adversely impact performance in keypoint estimation. As a result, we exclude these schemes in our evaluation of these models.}

\end{itemize}
\rbt{These modifications enable segmentation-based models to perform interactive keypoint estimation.}

\section{\rbttwo{Comparison with label-efficient methods}}\label{appen_sec:labelefficient}
\rbttwo{
Label-efficient methods primarily aim to improve model performance in scenarios with limited labeled data. These approaches mitigate data scarcity issues through strategies such as semi-supervised learning, one-shot learning, and active learning. For example, Jin et al.~\citep{jin2024rethinking} leverage semi-supervised learning with pseudo-labeling to enhance model accuracy given a few labeled data. Yao et al.~\citep{yao2021one} propose a one-shot keypoint detection approach that enables learning from a single labeled sample. Additionally, active learning-based methods optimize annotation efforts by selecting the most informative samples for labeling~\citep{shukla2022vl4pose}.}

\rbttwo{However, our approach addresses a fundamentally different challenge. Unlike label-efficient methods, we do not assume a scarcity of labeled data during training. Instead, our focus is on correcting inevitable prediction errors at inference time, which persist regardless of the quantity of labeled training data. Our method provides an efficient mechanism to refine these errors during inference, significantly reducing manual correction costs while maintaining high accuracy.}

\rbttwo{Even large-scale foundation models or other label-efficient methods cannot completely eliminate incorrect predictions. Even models trained on extensive datasets may still fail to ensure sufficient accuracy~\citep{kirillov2023segment_SAM_segment_anything}. This issue is particularly significant in medical applications, where precision is paramount. Inaccurate predictions can have severe consequences, necessitating human verification and correction. In such cases, completely eliminating human involvement is neither feasible nor desirable. Instead, our work focuses on making user interventions significantly more efficient by propagating corrections across multiple keypoints using a single user-provided keypoint. This reduces the annotation burden for medical professionals, ensuring both accuracy and efficiency.}

\rbttwo{In summary, while label-efficient methods aim to enhance model performance given limited labeled data, our interactive keypoint estimation framework addresses the challenge of efficiently refining unavoidable model errors at inference time, which is fundamentally distinct and complementary.}

\section{AI-assisted tool for growth potential estimation}\label{appen_sec:demo}
\setcounter{figure}{0}
\setcounter{table}{0}
In Fig.~\ref{supple_fig:demo}, we showcase a screenshot of our AI-assisted tool, underlining its practical usage in orthodontic and orthopedic planning. This tool, designed for ease of use and precision, enables clinicians to upload lateral cephalometric radiographs, perform annotations, and obtain accurate growth potential estimations efficiently. The integration of our deep learning model ensures immediate and simultaneous analysis, significantly aiding clinical decision-making. For more details, see the supplementary video.

\begin{figure*}[!b]
\begin{center}
  \includegraphics[width=0.95
\linewidth]{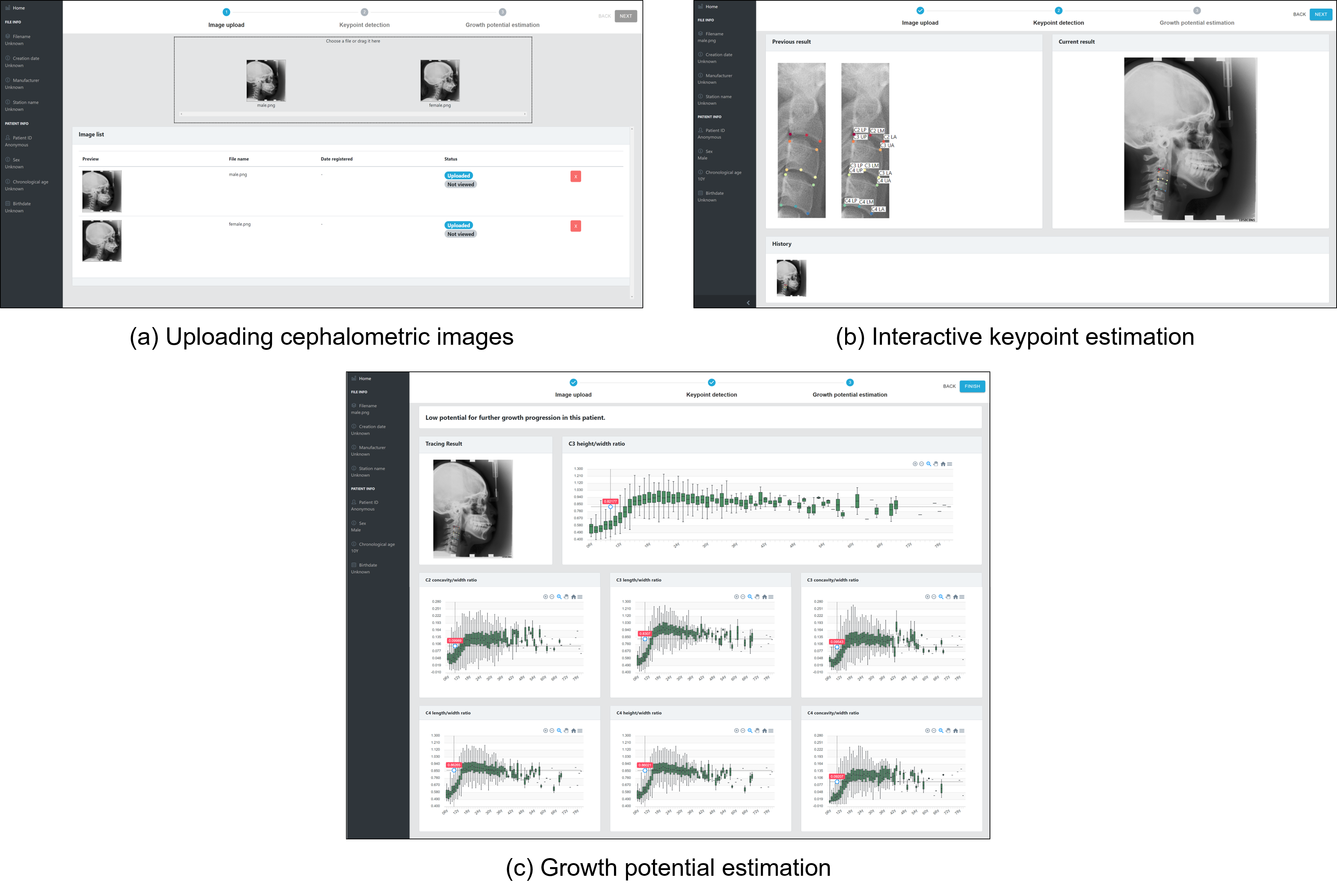}
\end{center}
\vspace{-15pt}
\caption{\rbt{\textbf{A screenshot of our AI-assisted growth potential estimation tool, showcasing its interface and functionality.} This tool enables clinicians to (a) upload lateral cephalometric radiographs, (b) perform annotations, and (c) obtain growth potential estimations. This highlights its utility in facilitating orthodontic and orthopedic planning with enhanced precision and ease of use.
}}
\label{supple_fig:demo}
\end{figure*}

\section{Qualitative comparison with the baseline models}\label{appen_sec:qual}
\setcounter{figure}{0}
\setcounter{table}{0}
In Fig.~\ref{suppe_fig:qual}, we present additional qualitative results comparing the predictions of \ourshort-v1 with other baseline models. Segmentation-based models demonstrate notable limitations, as user modifications often affect only nearby keypoints, and model revisions can degrade initial accuracy.
In contrast, ARNet-v1 refines distant inaccurate keypoints more effectively, reducing initial errors and propagating user feedback to correct distant mispredictions. These results emphasize the need for a tailored approach to interactive keypoint estimation, highlighting the proficiency of our proposed method in addressing these challenges.

\begin{figure*}[t]
\begin{center}
\includegraphics[width=1.0\linewidth]{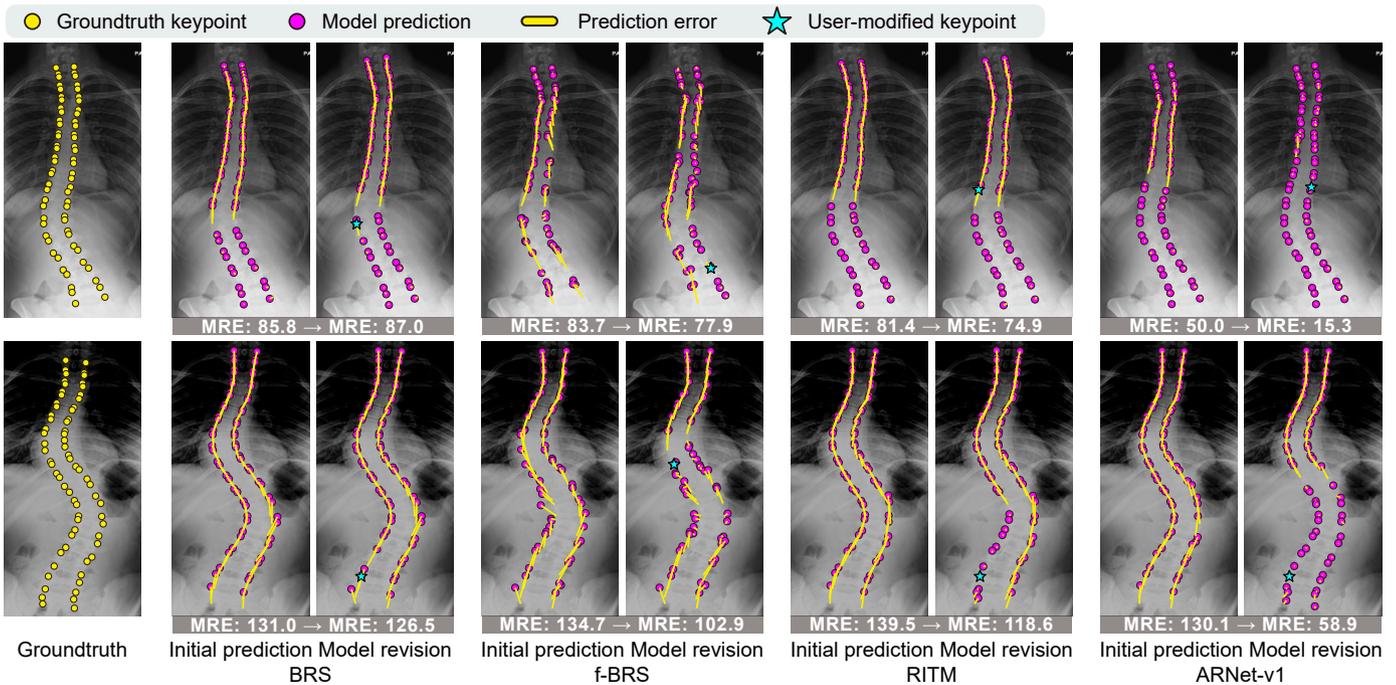}
\end{center}
\vspace{-10pt}
\caption{\rbt{\textbf{Qualitative comparison of \ourshort-v1 and the baseline models on the \spinewebdata dataset.} 
Initial prediction errors and errors after a single user modification are compared for each model. Prediction errors are visualized as lines connecting each predicted keypoint to its corresponding ground truth location. The length of these lines represents the magnitude of the prediction error: shorter lines indicate lower errors, while longer lines reflect greater errors.
}}
\label{suppe_fig:qual}
\end{figure*}

\section{Complete results of peak growth rates in cervical vertebrae}\label{appen_sec:diff}
\setcounter{figure}{0}
\setcounter{table}{0}
In Figs.~\ref{suppe_fig:diff1} and \ref{suppe_fig:diff2}, we present the complete set of standard growth curves and growth rates for all seven CVM features. They serve as a valuable reference for assessing a patient's remaining growth potential in clinical settings.

\begin{figure*}[ht]
\centering
\begin{center}
  \includegraphics[width=0.85\linewidth]{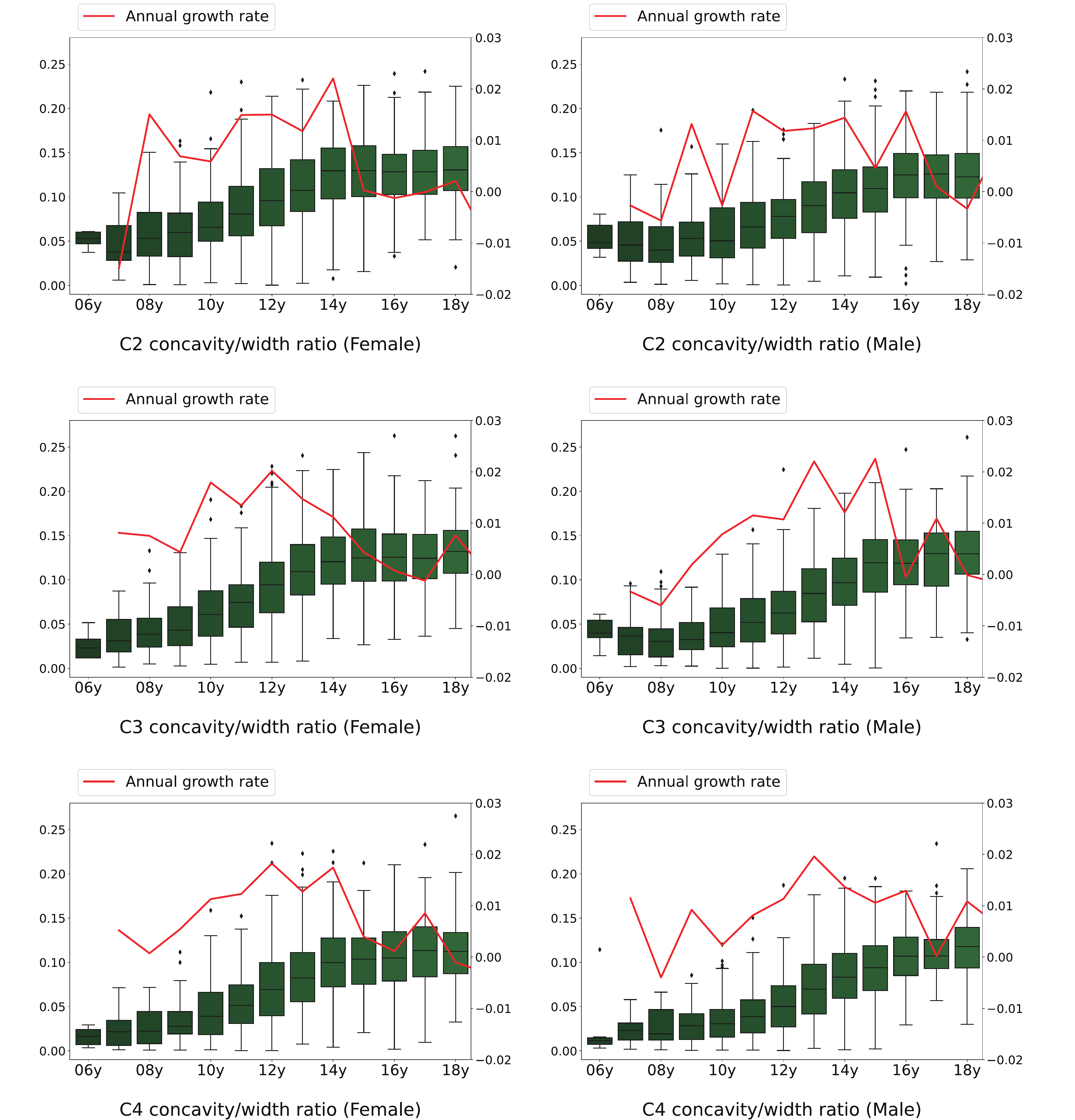}
\end{center}
   \caption{\textbf{Comprehensive annual growth rates based on standard growth curves of concavity/width ratio, demonstrating the growth peak for each cervical vertebra.} \rbtthree{The left y-axis indicates the CVM feature value, while the right y-axis represents the annual growth rate.}}
\label{suppe_fig:diff1}
\end{figure*}

\begin{figure*}[ht]
\begin{center}
  \includegraphics[width=0.85\linewidth]{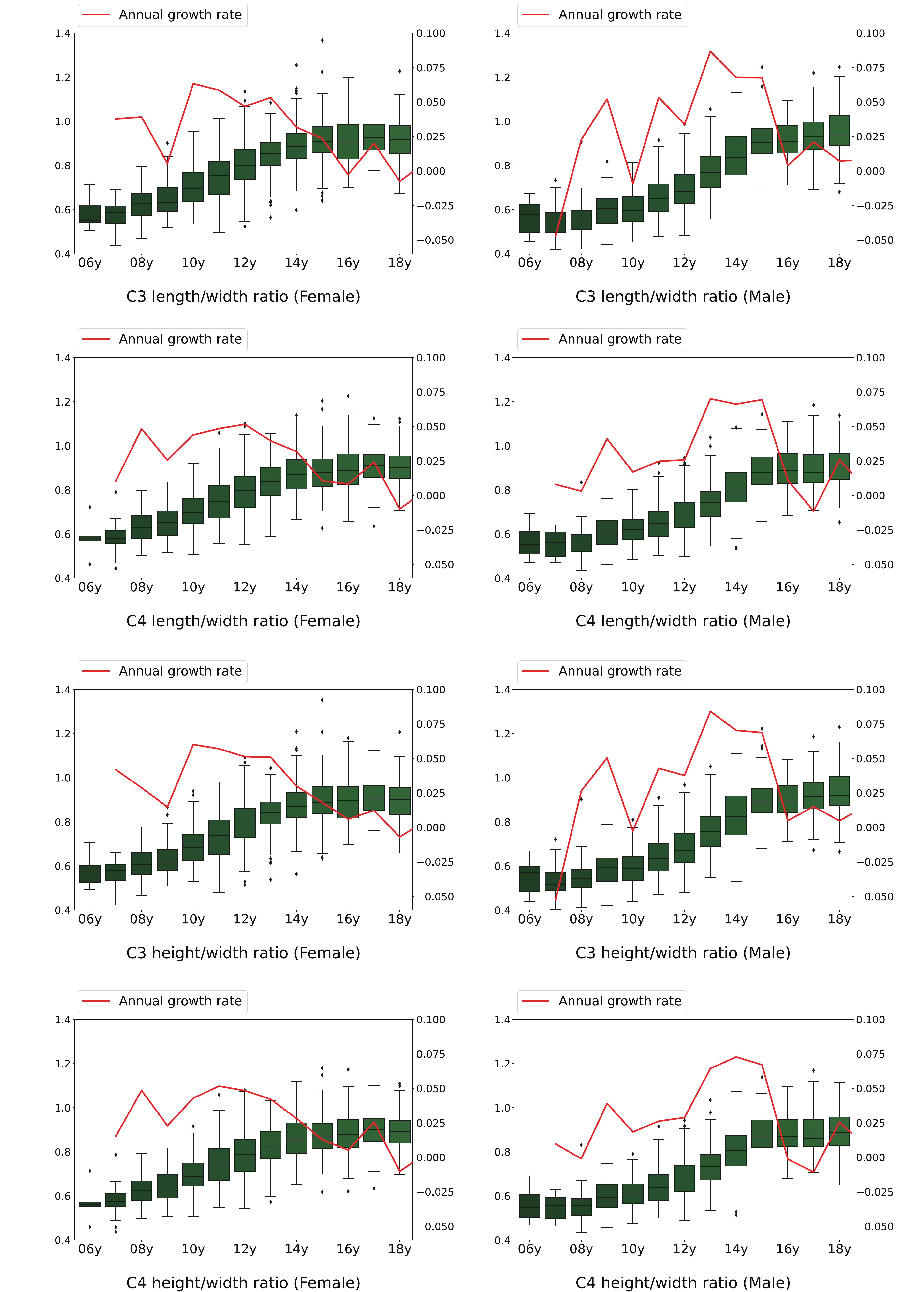}
\end{center}
   \caption{\textbf{Comprehensive annual growth rates based on standard growth curves of length/width and height/width ratios, demonstrating the growth peak for each cervical vertebra.} \rbtthree{The left y-axis indicates the CVM feature value, while the right y-axis represents the annual growth rate.}}
\label{suppe_fig:diff2}
\end{figure*}

\end{document}